\documentclass[11pt]{article}
\usepackage[margin=1in]{geometry}
\usepackage[T1]{fontenc}
\usepackage[utf8]{inputenc}
\usepackage{cite}
\usepackage{amsmath,amssymb,amsfonts}
\usepackage{algorithmic}
\usepackage{graphicx}
\usepackage{textcomp}
\usepackage{multirow}
\usepackage[font=small]{subfig}
\usepackage{xcolor}
\usepackage[hidelinks]{hyperref}
\usepackage{bm}

\newcommand{\change}[1]{#1}
\newcommand{\changes}[1]{#1}

\DeclareRobustCommand{\CoLoRA}{CoLoRA}

\def\BibTeX{{\rm B\kern-.05em{\sc i\kern-.025em b}\kern-.08em
    T\kern-.1667em\lower.7ex\hbox{E}\kern-.125emX}}

\begin{document}
\title{CoLoRA: Parameter-Efficient Fine-Tuning for Convolutional Models\\
A Case Study on Optical Coherence Tomography Classification}
\author{Mariano Rivera \qquad Angello Hoyos\\
\small Centro de Investigación en Matemáticas, A.C. (CIMAT)\\
\small Guanajuato, Guanajuato 36023, México\\
\small Corresponding author: \texttt{mrivera@cimat.mx}}
\date{}

\maketitle

\begin{abstract}
We introduce \textbf{CoLoRA} (Convolutional Low-Rank Adaptation), a parameter-efficient fine-tuning method for convolutional neural networks (CNNs). CoLoRA extends LoRA to convolutional layers by decomposing kernel updates into lightweight depthwise and pointwise components. This design reduces the number of trainable convolutional-update parameters by over 80\% compared with full convolutional fine-tuning, while allowing the learned updates to be merged into the pretrained convolutional kernels, thereby preserving the original model size and inference complexity. Experiments on MedMNIST datasets, particularly OCTMNISTv2, demonstrate that CoLoRA applied to VGG16 and ResNet50 achieves competitive classification performance while substantially reducing the number of trainable parameters. Comparisons with transfer learning, adapters, BitFit, and convolutional LoRA variants further characterize the trade-offs among predictive performance, trainable parameters, and training cost. Additional experiments on CIFAR-100 and ImageNet--R provide preliminary evidence that the proposed adaptation strategy also transfers to non-medical image-classification tasks. Peak GPU-memory measurements further show that parameter efficiency does not translate directly into proportional training-memory savings, with memory consumption depending strongly on the placement of the adapted convolutional layers. Overall, CoLoRA provides a parameter-efficient and deployment-efficient alternative to full fine-tuning for convolutional models.

\end{abstract}

\vspace{1em}

\section{Introduction}
\label{sec:intro}

Neural network models have grown dramatically in complexity and scale, particularly with foundational large language models (LLMs) \cite{zhou2024comprehensive, myers2024foundation}. Such models often have parameter counts ranging from tens of billions to trillions; for instance, GPT-4 (Generative Pre-trained Transformer) has approximately 1.5 trillion parameters, LLaMA 3.2 (Large Language Model Meta AI) has 90 billion parameters, and DeepSeek-V3 comprises 671 billion parameters \cite{naveed2023comprehensive}. 
This evolution underscores the need for efficient fine-tuning strategies that adapt pretrained models to downstream tasks without retraining the entire network. \emph{Parameter-Efficient Fine-Tuning} (PEFT) methods address this challenge by freezing most pretrained parameters and optimizing only a relatively small subset of weights or additional modules, showing success in both vision and language \cite{han2024parameterefficient,pmlr-v97-houlsby19a,pfeiffer2021adapterfusion}. In medical imaging, adapter-based strategies also facilitate fine-tuning of transformer architectures such as SAM \cite{Kirillov_2023_ICCV,WU2025103547}.
The Low--Rank Adaptation (LoRA) method, notable for simplicity, is designed to efficiently fine-tune large pre-trained models for specific tasks \cite{hu2022lora}. This is achieved by introducing low-rank learnable matrices into the model, allowing most of the original parameters to remain frozen. This significantly reduces the computational and storage costs associated with full fine-tuning while maintaining high task performance. Recent work in intelligent rehabilitation has also shown the value of deep models that preserve spatial structure. Xing \emph{et al.} used 3-D graph deep learning with laser point-cloud data for hand segmentation in rehabilitation scenarios, and later proposed a 3-D graph-based method for non-contact gesture recognition in human--computer interactive rehabilitation \cite{xing2025visual,xing2026rehabilitation}. Although these methods are not PEFT approaches, they reinforce the importance of efficient deep-learning models that capture local geometric or spatial relationships in medical and assistive applications.

Structure-aware modeling is also relevant in other engineering disciplines when spatial irregularity or asymmetry affects system behavior. For example, Burgan analyzed the seismic response of an asymmetric L-shaped building under different loading conditions and structural configurations\cite{burgan2021numerical}. Although this setting is not directly related to neural-network fine-tuning, it illustrates the broader importance of accounting for structural organization when modeling spatially dependent systems. In CoLoRA, this motivation appears at the kernel level, where spatial and cross-channel relationships are modeled through structured depthwise and pointwise updates.

\begin{figure}[!ht]
\centering
\includegraphics[width=0.9\linewidth]{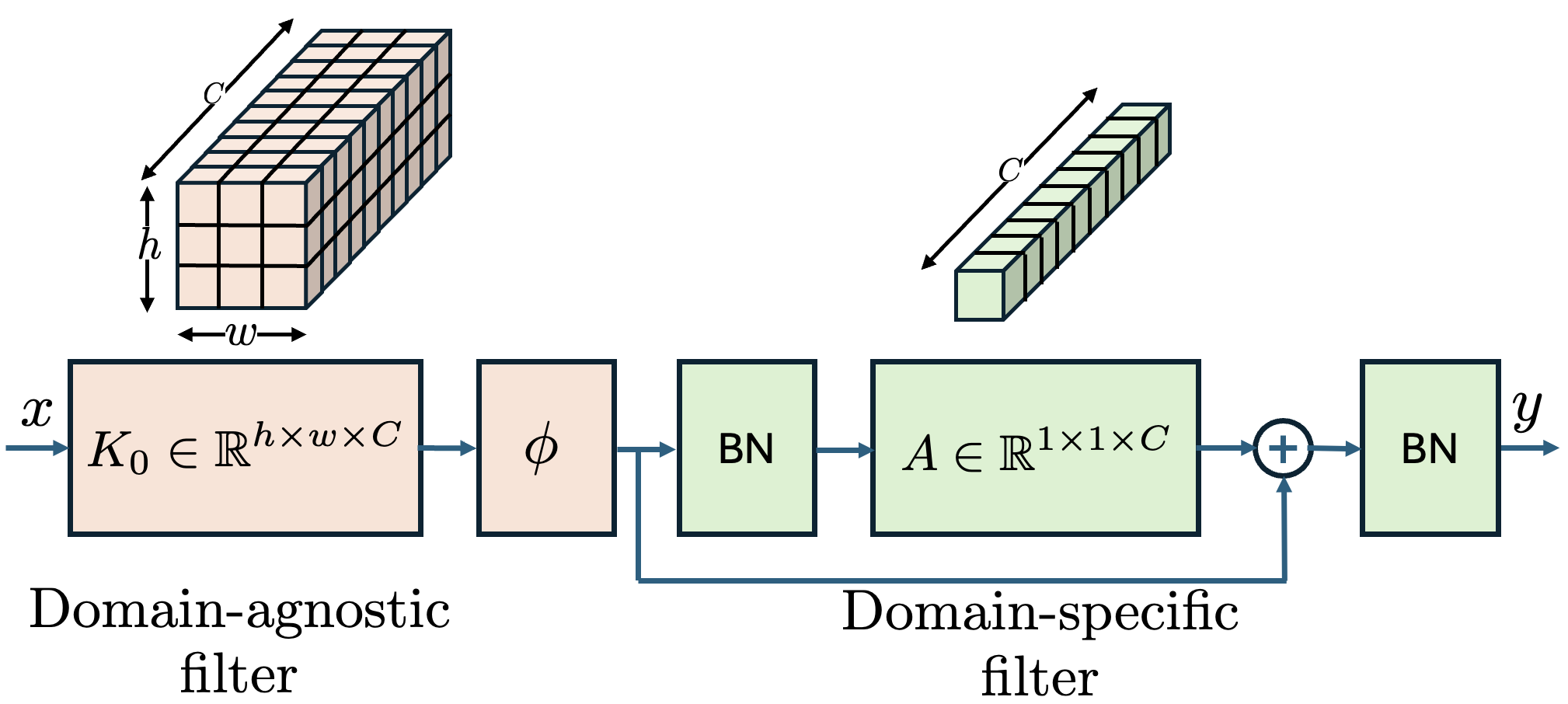}
\caption{Schematic of Rebuffi's CNN adapter. A $1 \times 1$ adapter after each convolutional block mixes preceding activations for domain-specific adaptation while freezing backbone filters.}
\label{fig:cnn_adapter}
\end{figure}

Nevertheless, PEFT in medical imaging remains comparatively underexplored \cite{dutt2024parameterefficient}, where CNN-based transfer learning is still prevalent \cite{kim2022transfer,salehi2023study}. Although CNNs are smaller than LLMs, effective fine-tuning remains crucial. The main challenges include overfitting, feature misalignment, and inefficient parameter adaptation—issues exacerbated when trainable parameters are numerous relative to available data \cite{tajbakhsh2016convolutional,kornblith2019better}. Cross-domain transfer can suffer when source and target distributions differ substantially, and careful hyperparameter choices (learning rate, batch size, freezing policy) are needed to avoid catastrophic forgetting \cite{kolesnikov2019large}. In his seminal work, Rebuffi \emph{et al.} proposed CNN adapters—residual modules that enable domain-specific adaptation while freezing the backbone \cite{rebuffi2017learning,rebuffi2018efficient}. Fig.~\ref{fig:cnn_adapter} illustrates how the trainable adapter (in light green) is introduced after a frozen pretrained convolutional layer (in light orange). As one can note, Rebuffi's adapter permanently modifies the architecture and increases the inference cost. 

To address this, we propose \textbf{CoLoRA}, which extends LoRA to convolutional architectures. While LoRA updates dense layer weights by factorizing them into two low-rank matrices (see Fig. \ref{fig:lora_conv_sconv}(a)), our method updates convolutional kernels (Fig. \ref{fig:lora_conv_sconv}(b)) by decomposing them into pointwise ($1 \times 1$) and depthwise convolutions (with spatial dimensions but only one channel); see Fig. \ref{fig:lora_conv_sconv}(c).

We implement CoLoRA on VGG16 \cite{simonyan2014very} and ResNet50 \cite{he2016deep,he2016identity}, both ImageNet-pretrained. Following the computing order of Inception-Net \cite{Szegedy_2015_CVPR}, we first capture cross-channel correlations to reduce tensor dimensionality, then model spatial correlations in the reduced representation. Our CoLoRA layer can straightforwardly be generalized for fine-tuning convolutional layers in other dimensions than two; provided that we have a ready implementation of the depthwise convolution layer. The strategy, inspired by Inception Net \cite{Szegedy_2015_CVPR}, is the first to look for cross-channel correlations to reduce the dimension of the tensors along the channels. Next, the model looks for spatial correlations in a lower-dimension tensor (a tensor with a single channel). This order of operations is computationally efficient and significantly reduces the number of parameters required for fine-tuning compared to traditional methods. Furthermore, depending on layer placement, CoLoRA can provide a favorable trade-off among predictive performance, trainable parameters, and training cost.

\change{The central contribution of CoLoRA is a convolution-specific structured PEFT formulation combining pointwise–depthwise kernel factorization, exact fusion into the pretrained convolutional weights, and an optional merge-reset mechanism that allows structured updates to accumulate during training.}

We evaluated on \textbf{OCTMNISTv2} \cite{medmnistv1,medmnistv2}---a clinically relevant, class-imbalanced benchmark (ChN, DME, Drusen, Normal) that remains challenging even for recent architectures \cite{manzari2023medvit,manzari2025medical,yang2025medkan}. CoLoRA-based CNNs obtain a best test accuracy of 0.966 in the VGG16 placement study and an AUC of 0.995 in the distilled VGG16 experiment, while preserving the original inference graph after merging.

The remainder of this paper is structured as follows. Section \ref{sec:background} reviews related work, Section \ref{sec:proposal} details the proposed CoLoRA method, Section \ref{sec:experiments} presents experimental validations and results, Section \ref{sec:conclusions} gives the conclusions, and Section \ref{sec:future} outlines future research directions.

The source code implementing CoLoRA is publicly available at
\url{https://github.com/ajhoyos/CoLoRA}.

\section{Related Methods}
\label{sec:background}

\subsection{Convolutional Layers}
\label{ssec:convolutional}

The Convolutional Layer is the fundamental building block of Convolutional Neural Networks (CNNs) designed to process grid-like data such as images \cite{simonyan2014very, he2016deep,he2016identity, venkatesan2017convolutional}. 
A convolutional layer processes grid-structured data with learnable filters, extracting hierarchical spatial features.
Convolutional layers are used in image classification, object detection, segmentation, and more. They are key to achieving state-of-the-art performance in tasks involving visual data. It performs a convolution operation by applying a set of learnable filters (kernels) to input data, extracting spatial hierarchies of features such as edges, textures, and shapes. 

Let $x\!\in\!\mathbb{R}^{H \times W \times C}$ be the data to process and $K^t\!\in\!\mathbb{R}^{h \times w \times C}$ ($t=1,\dots,T$) be the kernel filters, then the output $y\!\in\!\mathbb{R}^{H \times W \times T}$ is computed with
\begin{align}
   y(i,j,t) &= (K^t \circledast x)(i,j) \nonumber \\
            &= \sum_{l,m,k} K^t(l,m,k)\, x(i+l, j+m, k).
\end{align}
where $ \circledast$ denotes the correlation operation, $(i, j)$ the spatial position, and $(l, m, k)$ index the kernel/input coordinates, and $t$ indexes the output filter. The filters operate on small local regions of the input image, allowing the network to capture spatial relationships. Each filter is applied across the entire input, significantly reducing the number of parameters. Fig.~\ref{fig:lora_conv_sconv}b depicts a 2D-convolutional filter or kernel with dimension $(h \times w \times C)$; \emph{i.e}, \emph{h}eight, \emph{w}idth and \emph{c}hannels. Although the filter has dimensions, it is applied by shifting the kernel only in the spatial coordinates, whereas the kernel depth $C$ is equal to the number of channels in the input $x$. In addition, multiple filters extract different features, allowing rich representations of the data. The most relevant hyper-parameters of the convolutional layers are: the \emph{number of filters} that specify the number of unique features to extract, the \emph{filter size} that determines the receptive field of the filter (\emph{e.g.}, $3 \times 3$, $5 \times 5$), the \emph{stride} that defines the step size of the filter spatial displacement, and the \emph{padding} that defines the added borders to the input to control the spatial dimensions of the output.

\subsection{CNN Adapters}

The CNN adapters proposed by Rebuffi introduce residual $1\times1$ trainable convolutional layers while freezing the original filters \cite{rebuffi2017learning,rebuffi2018efficient}; see Fig.~\ref{fig:cnn_adapter}. Let $x$ be the output of the pretrained convolutional layer (including the activation $\phi$). Then, the Rebuffi adapter is computed as follows:
\begin{align}
    x'  & \leftarrow \textrm{BN} (x), \\
    x'(i,j,t) &\leftarrow \sum_{k=1}^{C} A_{k,t}\, \phi\!\big((K^t \circledast x')(i,j)\big), \\
    x &\leftarrow x+ x', \\
    y &\leftarrow \phi\left(\textrm{BN}(x) \right)
\end{align}
where $\textrm{BN}$ is a batch-normalization layer \cite{ioffe2015batch}, $A\in\mathbb{R}^{C \times T}$ is a $1 \times 1 \times C$ kernel and $\phi$ an activation. Since the feature subspace spanned by frozen filters is unchanged, features absent from the backbone cannot be recovered. Chen et al. \cite{chen2023chapter} combines CNN adapters with transformer-style Houlsby adapters \cite{pmlr-v97-houlsby19a}, improving flexibility but increasing inference cost.

\subsection{LoRA: Low-Rank Adaptation}
\label{ssec:lora}

LoRA is a particular case of a general family of adapters introduced in the context of transformers \cite{hu2022lora,hetowards,steitz2024adapters}.

\begin{figure}[!ht]
\centering
\includegraphics[width=1\linewidth]{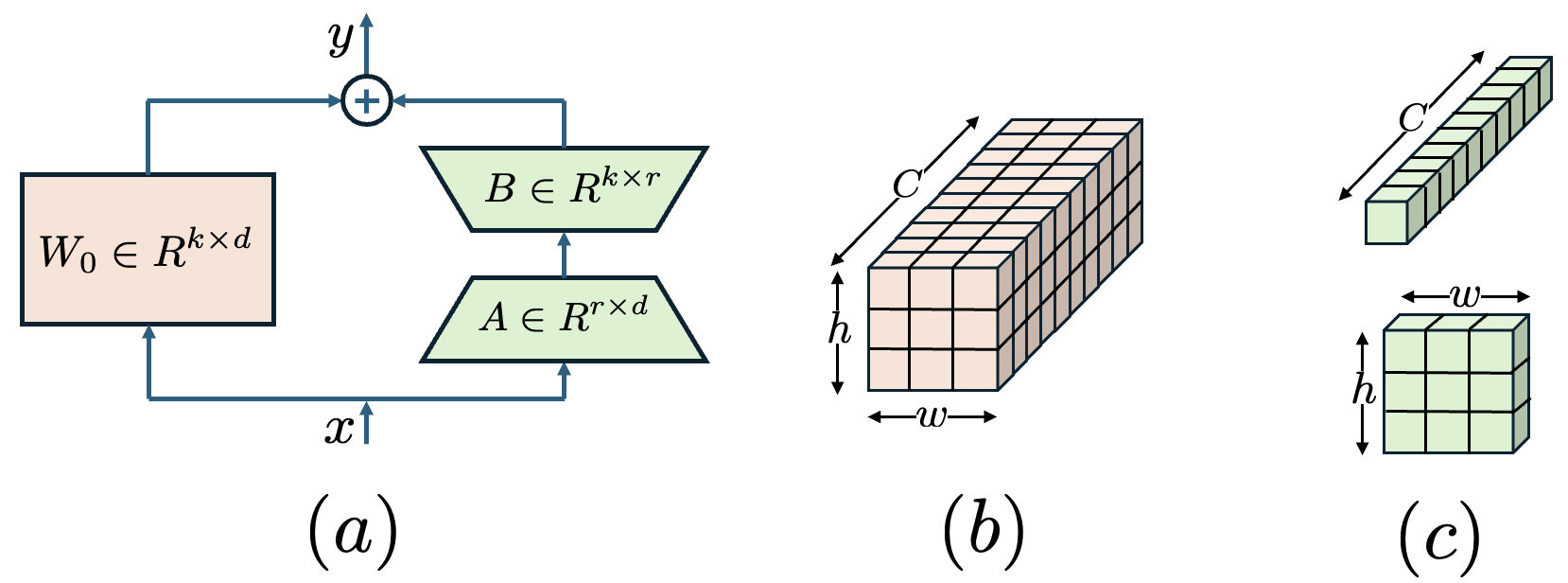}
\caption{(a) LoRA on dense layers. (b) 2D convolutional kernel. (c) depthwise–pointwise factorization of a 2D convolution.}
\label{fig:lora_conv_sconv}
\end{figure}

LoRA fine-tunes large models by adding low-rank updates while freezing backbone weights. Assume the pretrained linear layer $y=W_0\,x$, where $W_0$ is the matrix of weights and $\Delta W$ its update required by the fine-tuning process. Then, LoRA constrains the update to be a low-rank matrix that can be factorized by $\Delta W = B\,A$ with $A \in \mathbb{R}^{r \times d}$ and $B\in\mathbb{R}^{k \times r}$, $r \ll \min(d,k)$.  Here, $d$ and $k$ are the dimensions of the original weight matrix, and $r$ is the rank, which is typically set to a small value. Hence
\begin{equation}
    y \;=\; (W_0 + \Delta W)\,x \;=\; W_0x + B \, Ax.
\end{equation}
 In LoRA the fine-tuning is achieved by freezing the pre-trained weights $W_0$ and training the Low-Rank Matrices $A$ and $B$; setting initially $A \sim \mathcal{N}(0,\sigma^2)$ and $B=0$. Fig.~\ref{fig:lora_conv_sconv}(a) illustrates LoRA. Advantages include
 \begin{itemize}
    \item \textbf{Efficiency:} By keeping most pretrained parameters fixed, LoRA reduces the number of trainable parameters and associated gradient and optimizer-state storage, although peak training-memory consumption also depends on activation storage and the adapted layers.
    \item \textbf{Modularity:} The low-rank updates can be stored and applied separately, making the method highly modular.
    \item \textbf{Scalability:} LoRA is suitable for fine-tuning very large models without retraining or duplicating the entire model.
\end{itemize}

\noindent LoRA is widely used for fine-tuning foundational models in the domains of natural language processing (NLP), computer vision, and other domains where fine-tuning large models is computationally expensive \cite{zhou2024comprehensive,myers2024foundation,awais2025foundation}. In particular, in Transformers-based architectures where the attention mechanism is based on dense matrices (in general denoted by $Q$, $K$, and $V$) with dimensions $d=k$ and $d$ large; \emph{e.g.}, $d=1024$. Its modular nature also makes it compatible with several architectures and training regimes.

Recent ophthalmic studies further demonstrate the applicability of low-rank fine-tuning in medical imaging. Lee et al. compared classifier-only training, partial fine-tuning, and LoRA across four vision foundation models for carotid atherosclerosis prediction from retinal fundus images. Among the evaluated configurations, DINOv2 with LoRA provided the best overall combination of classification performance, prognostic utility, and vessel-aligned explainability \cite{lee2025retinal}. Zhen et al. combined two pretrained vision transformers with deep canonical correlation analysis for multimodal color-fundus and OCT-based age-related macular degeneration classification. Their LoRA-based adaptation required approximately
0.49\% of the trainable parameters of full fine-tuning while achieving an F1-score of 0.948, an AUC-ROC of 0.991, and an accuracy of 0.949 \cite{zhen2026lowrank}. These studies support the effectiveness of low-rank adaptation in ophthalmic imaging; however, they focus primarily on transformer-based encoders and do not formulate a convolution-specific, mergeable update for CNN kernels. CoLoRA addresses this complementary setting by explicitly preserving the spatial and channel structure of convolutional operators.

\subsection{Separable Convolutional Layers}
\label{ssec:cseparable}

Depthwise separable convolutions, central in InceptionNet \cite{Szegedy_2015_CVPR} and MobileNet models \cite{howard2017mobilenets}, factorize a standard convolution into depthwise and pointwise steps:
\begin{align}
    z(i,j,k) &= (K_d \circledast x)(i,j,k) \nonumber \\
             &= \sum_{i',j'} K_d(i',j')\, x(i+i',j+j',k), \\
    y(i,j)   &= (K_p \circledast z)(i,j) \;=\; \sum_{k} K_p(k)\, z(i,j,k),
\end{align}
which we summarize as $y = K_p \circledast (K_d \circledast x)$. This reduces parameters and FLOPs. Fig.~\ref{fig:lora_conv_sconv}(b) illustrates a two-dimensional convolutional kernel and Fig.~\ref{fig:lora_conv_sconv}(c) a $1 \times 1 \times 1$ pointwise kernel (top) and a $3 \times 3 \times 1$ depthwise kernel (bottom).

\section{Proposed Method}
\label{sec:proposal}
Pre-trained backbones trained on unrelated datasets are employed for developing domain-specific models. The standard training procedure involves two stages. First, the backbone acts as a fixed universal feature extractor, transforming input data into a feature space suitable for training subsequent domain-specific layers. The second stage fine-tunes a subset of weights within the backbone alongside the newly added layers, further improving model accuracy.

\subsection{Convolutional Low-Rank Adaptation}

\begin{figure}[t!]
\centering
\includegraphics[width=0.75\linewidth]{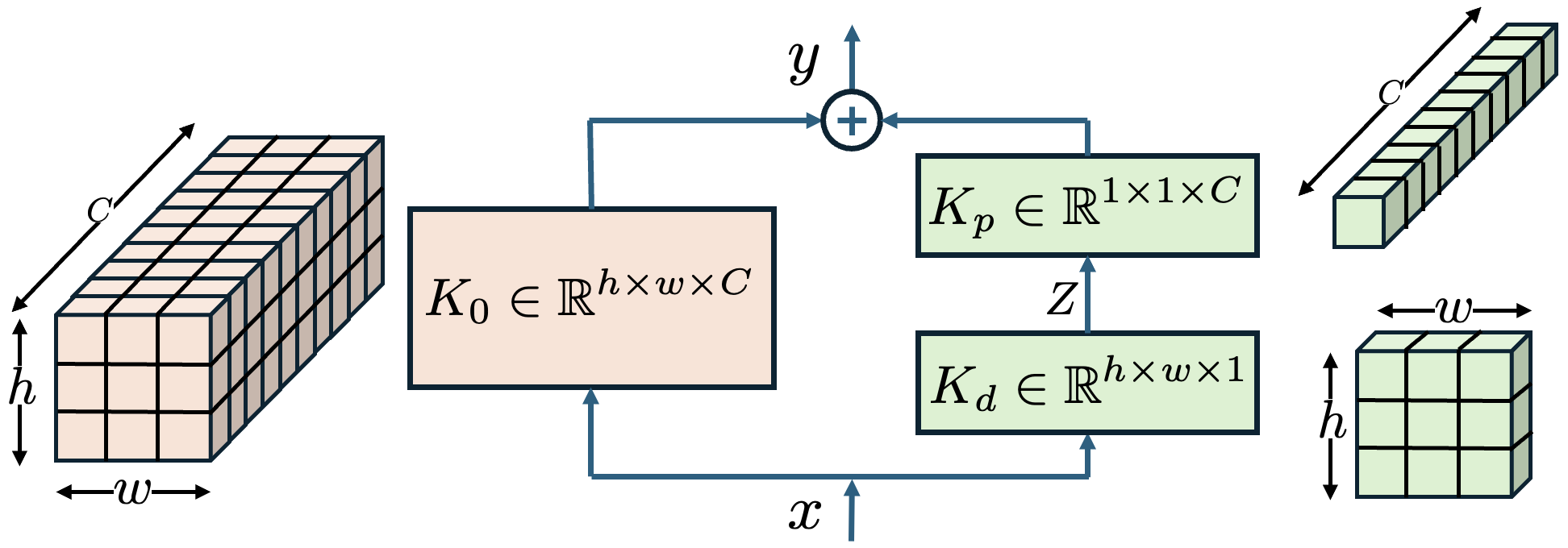}
\caption{CoLoRA layer: a trainable depthwise--pointwise residual is added to a frozen convolution; learned updates can be merged during training and are fused into the backbone before inference, preserving the original inference graph.}
\label{fig:colora}
\end{figure}

CoLoRA is a convolutional extension of the LoRA paradigm. While LoRA constrains the update weights (matrix) $\delta K$  by the product of two low-rank matrices, CoLoRA approximates the kernel update $\delta K$ by applying separable convolutions. Particularly, CoLoRA constrains the updates to be expressed with a learnable depthwise separable convolution and keeps the pre-trained weights frozen. This significantly reduces the number of learnable parameters, especially when the number of input channels and filters is large. We observed that this strategy results in a stable and accurate fine-tuning procedure. Given a pretrained layer 
\begin{equation}
    y = K_0 \circledast x;
\end{equation}
then, CoLoRA fine-tuning consists of updating the Kernel $K_0 \in \mathbb{R}^{h \times w \times C}$ with $\Delta K$ such that
\begin{align}
    y &= (K_0 + \Delta K_0)\circledast x \nonumber \\
      &= K_0 \circledast x + \Delta K_0 \circledast x.
\end{align}
In CoLoRA, we constrain the kernel update $\Delta K$ to be a separable kernel
\begin{equation}
    \label{eq:KpKd}
    \Delta K_0 = K_p \circledast K_d;
\end{equation}
where $K_p$ and $K_d$ denote the pointwise and depthwise kernels, respectively (see Fig.~\ref{fig:colora}).  The spatial depthwise layer is created preserving the following properties of the pre-trained convolutional
layer: \emph{data type, number of filters, kernel size, stride, padding, data format, dilation rate, activation, use of bias}. Hence, we freeze $K_0$ and train only $(K_p,K_d)$, and the optional bias with $\Delta b$. We use  Glorot-uniform for $K_p$ and zeros for $K_d$ (and $\Delta b$) as initializers. After training, we merge the updates:
\begin{align}
    \label{eq:margeK0}
    K_0 &\leftarrow K_0 + K_p \circledast K_d, \qquad \\
    \label{eq:margeb0}
    b_0 &\leftarrow b_0 + \Delta b,
\end{align}
optionally at fixed intervals (e.g., after each epoch).

In more detail, assuming a kernel $K_0$ has size $h \times w \times C$. \change{Then the pointwise $K_p$ is of size $1 \times 1 \times C$ and the depthwise kernel $K_d$ is of size $h \times w \times 1$.} Then, taking advantage of the technique called broadcasting of the programming language Python and extended to any tensor library (TensorFlow or PyTorch), their composition produces the kernel $\Delta K_0$ with size $h \times w \times C$. In case of using other programming language, the broadcasting implementation implies simple loops.   

Moreover, cumulative updates correspond to
\begin{align}
    K_0 &\leftarrow K_0 + \sum_{e=1}^{E} K_p^{(e)} \circledast K_d^{(e)}, \qquad \\
    b_0 &\leftarrow b_0 + \sum_{e=1}^{E} \Delta b^{(e)}.
\end{align}
Thus, $\Delta K_0$ lies in the span of $\{K_d^{(e)}\}$, capturing diverse spatial patterns accumulated over training. 

\changes{
    For each output filter, if the spatial dimensions of an individual merged update are flattened, the update can be written as the outer product of a spatial vector derived from $K_d^{(e)}$ and a channel vector derived from $K_p^{(e)}$. Therefore, each individual merged update has matrix rank at most one under this unfolding, while the cumulative update after $E$ merge-reset stages has rank at most $\min(E,h w,C)$. This is an upper bound: the effective rank can be lower when successive components are linearly dependent. Consequently, CoLoRA does not introduce an explicit per-update rank hyperparameter $r$ as standard LoRA does, but the number and frequency of merge-reset stages can increase cumulative adaptation capacity. Merge frequency therefore acts as an indirect capacity-control mechanism and is not independent of rank-like capacity, although it is not equivalent to jointly optimizing a fixed-rank LoRA update because the components are learned sequentially after previous updates have already been fused into the backbone.
}

At inference time, the CoLoRA branch is not executed as an additional residual path. Instead, its learned contribution is fused into the backbone by replacing the original kernel and bias with the merged parameters above. Consequently, the deployed model keeps the same convolutional operators, tensor shapes, and forward graph as the original pretrained CNN, but with updated weights. In practical terms, CoLoRA is fused into the backbone through kernel-and-bias merging, so no extra CoLoRA module remains active during inference. After this merging step, the auxiliary CoLoRA layers are removed from the deployed model.

The benefits of using separable convolutions are their \textbf{efficiency}, significant reduction in computational complexity, and number of parameters; \textbf{flexibility}, limited computational resource requirements; and \textbf{performance}, competitive accuracy compared to standard convolutions. Our approach also has the advantages of the LoRA update scheme. 

\subsection{Generalization to Higher Dimensions}

Separable convolutions in 3D are not implemented in \emph{TensorFlow} or \emph{PyTorch} at the moment. Hence, here we present how to generalize the CoLoRA layer to other dimensions than two; \emph{i.e.,} to one or three dimensions. 

\begin{figure}[ht!]
\centering
\includegraphics[width=0.75\linewidth]{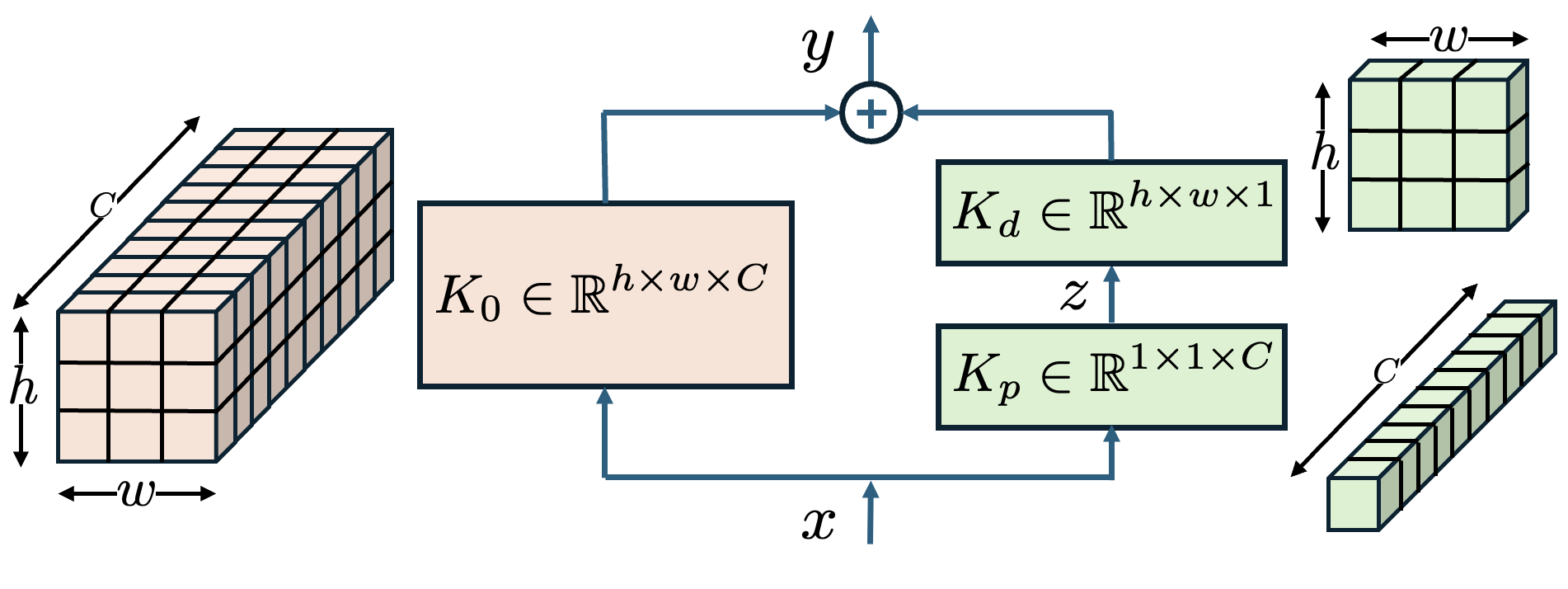}
\caption{Inception-style CoLoRA: pointwise mixing precedes depthwise spatial filtering.}
\label{fig:colora_inception}
\end{figure}

Because convolution is linear, we can swap the separable order:
\begin{equation}
    \Delta K \;=\; K_d \circledast K_p,
\end{equation}
first mixing channels (pointwise) and then applying spatial filtering (depthwise), as in Inception Net \cite{Szegedy_2015_CVPR, Chollet_2017_CVPR}; This generalizes naturally to 1D and 3D; see Fig.~\ref{fig:colora_inception}. 

\begin{figure}[ht!]
\centering
\includegraphics[width=1\linewidth]{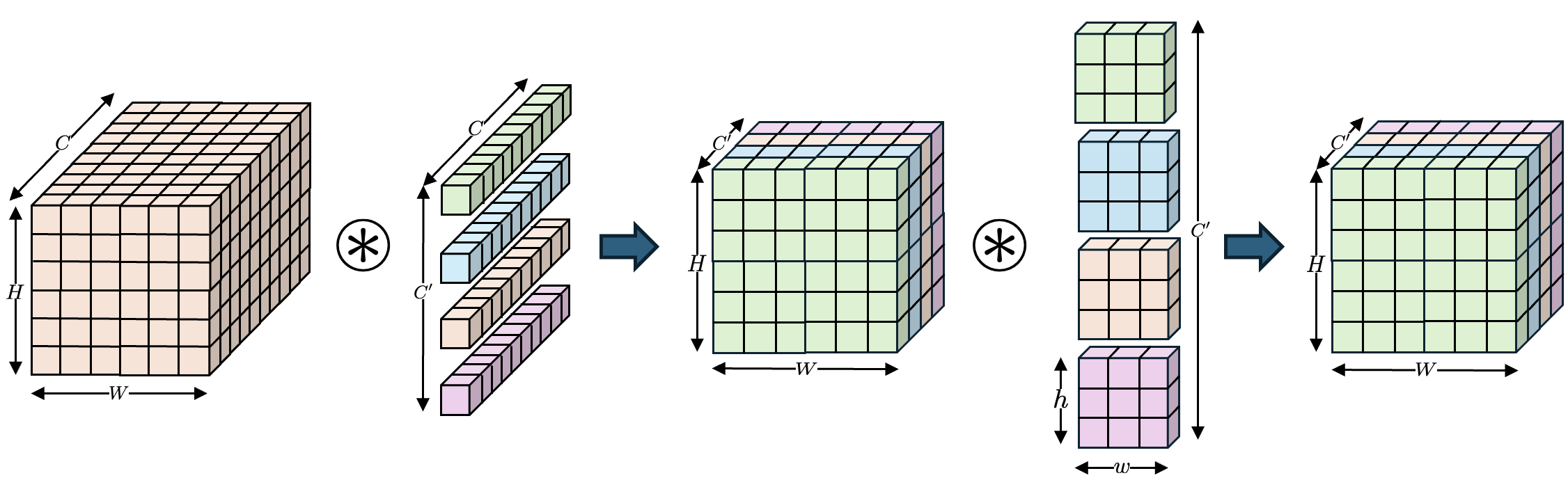}
\caption{Operations of the CoLoRA layer (2D): $1 \times 1$ pointwise mixing per location followed by per--channel depthwise spatial convolution.}
\label{fig:colora_inception_2}
\end{figure}

Fig.~\ref{fig:colora_inception_2} illustrates the 2D case. For the convolutional 3D layer case with input $x \in \mathbb{R}^{H \times W \times D \times C}$, each filter in the CoLoRA layer implements: 
\begin{align}
    \label{eq_colorav2}
    z_1 & = K_p \circledast x \\
    z_2 & = K_d \circledast z_1 \\
    z_3 &= K_0 \circledast x  \\
    y   & = z_3 + z_2
\end{align}
where $\circledast$ denotes the correlation operation, $K_p \in \mathbb{R}^{1 \times 1 \times 1 \times C}$ and 
$K_d \in \mathbb{R}^{h \times w \times d \times 1}$. Thus, if we apply the $C'$ filters, it results in $z_1, z_2, z_3, y \in \mathbb{R}^{H \times W \times D \times C'}$. This order of operations is computationally efficient: we computed $C'$ weighted sums of the input tensor channels and computed $C'$ 3D-convolutions of a single channel. Moreover, it results in a straightforward generalization of the CoLoRA layers for fine-tuning convolutional layers in other dimensions than two, provided that we have a ready implementation of the depthwise convolution layer. 

Table~\ref{tab:parameters} compares the number of trainable and inference-stage parameters for a convolutional layer with kernel size $h \times w$, $C$ input channels, and $T$ output filters. A standard convolution has $N_O=h\,w\,C\,T$ parameters. A CNN adapter in the style of Rebuffi's adapter trains a $1 \times 1$ residual convolution plus batch-normalization parameters, but this extra branch remains active at inference. CoLoRA trains a depthwise--pointwise update with $N_C=h\,w\,T+C\,T$ parameters and then merges this update into the original convolutional kernel. Thus, after merging, CoLoRA does not increase the inference-stage parameter count or forward graph with respect to the original model.

\begin{table}[ht!]
    \centering
    \begin{tabular}{|l|l|c|}
    \hline
        \textbf{Convolution} & \textbf{Training} & \textbf{Inference} \\
        \hline
        Original     & $N_O = h\,w\,C\,T$         & $N_O$ \\
        CNN Adapter  & $N_A = C\,T + 2$           & $N_O + N_A$ \\
        CoLoRA       & $N_C = h\,w\,T + C\,T$     & $N_O$ \\
        \hline
    \end{tabular}
    \caption{Trainable parameter counts for a convolutional layer under different strategies. $(h,w)$: kernel size; $C$: input channels; $T$: filters. The CNN adapter includes BN (2 trainable parameters).}
    \label{tab:parameters}
\end{table}

The parameter reduction follows directly from the separable constraint. The trainable-parameter ratio between CoLoRA and a full convolutional update is
$$
\frac{N_C}{N_O}=\frac{h\,w\,T+C\,T}{h\,w\,C\,T}=\frac{h\,w+C}{h\,w\,C}.
$$ 
For a common $3 \times 3$ convolution and moderate-to-large $C$, this ratio approaches $1/9$, corresponding to an approximate 89\% reduction in trainable update parameters. From the perspective of low-rank modeling, CoLoRA can be interpreted as a structured low-rank approximation of the unfolded convolutional update: the pointwise component captures cross-channel correlations, while the depthwise component captures local spatial patterns. Unlike standard LoRA, CoLoRA does not introduce an explicit per-update rank hyperparameter $r$; its per-update capacity is determined by the native convolutional dimensions and by the selected layers where CoLoRA is applied, while the cumulative capacity can also increase with the number of merge-reset stages, as discussed in Sec.~III-A.

\subsection{Implementation Details}

In our implementation, we selected as backbones two commonly used convolutional network architectures: VGG16 \cite{simonyan2014very} and ResNet50v2 \cite{he2016deep}; both pre-trained with ImageNet. VGG16 is widely used in transfer-learning implementations due to the simplicity of the architecture. ResNet50v2 is selected for  parameter reduction compared to VGG16 while preserving high performance. 

\begin{figure*}[ht!]
\centering
\includegraphics[width=\linewidth]{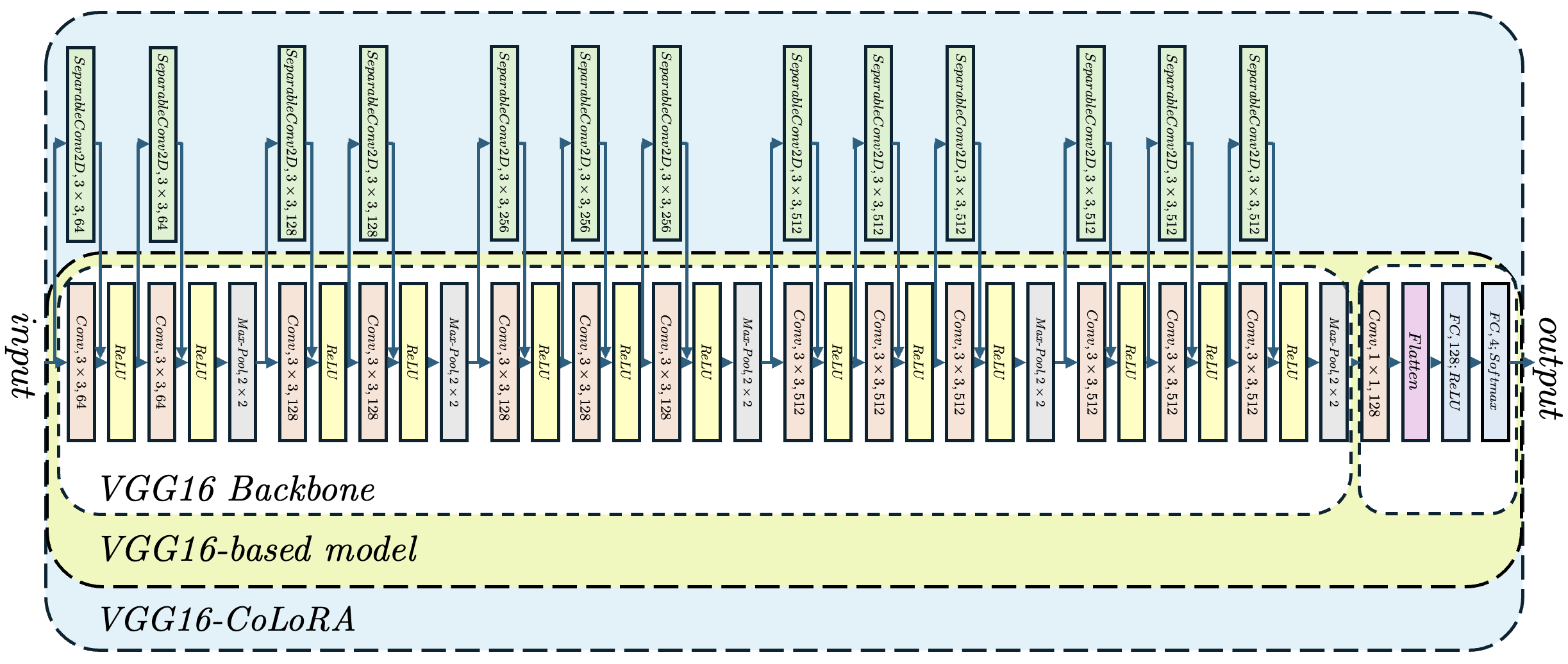}
\caption{VGG16–CoLoRA architecture. Each convolution is augmented with a CoLoRA residual; the backbone is initialized from ImageNet.}
\label{fig:vgg16_colora}
\end{figure*}

Due to its simplicity, we illustrate the VGG16--CoLoRA model in Fig.~\ref{fig:vgg16_colora}. We first built a VGG16-based model with explicit Conv2D and ReLU activation layers and initialized the backbone with ImageNet-pretrained weights. We then added a residual CoLoRA bypass to each selected Conv2D layer, preserving the number of filters, stride, dilation, padding, and related convolutional settings of the original layer. The $K_p$ kernel was initialized with Glorot-uniform values \cite{glorot2010understanding}, whereas the $K_d$ kernel was initialized to zero. For the task-specific decision stage, we used a $1 \times 1$ convolution to reduce the number of channels from 512 to 128, followed by two fully connected layers: a hidden linear layer with 128 units and ReLU activation, and a four-unit output layer without activation, corresponding to logits for the four classes. The VGG16--CoLoRA model has 15.58 million total parameters; only 2.54 million are trained by backpropagation, and the remaining weights are updated through the merging procedure after each epoch. We used Adam with its default settings. Experiments were performed on an NVIDIA RTX 3090 GPU, and each epoch took approximately 142 seconds.

\section{Experiments}
\label{sec:experiments}

\subsection{Dataset} 

\begin{figure}[!t]
\centering
\includegraphics[width=0.8\linewidth]{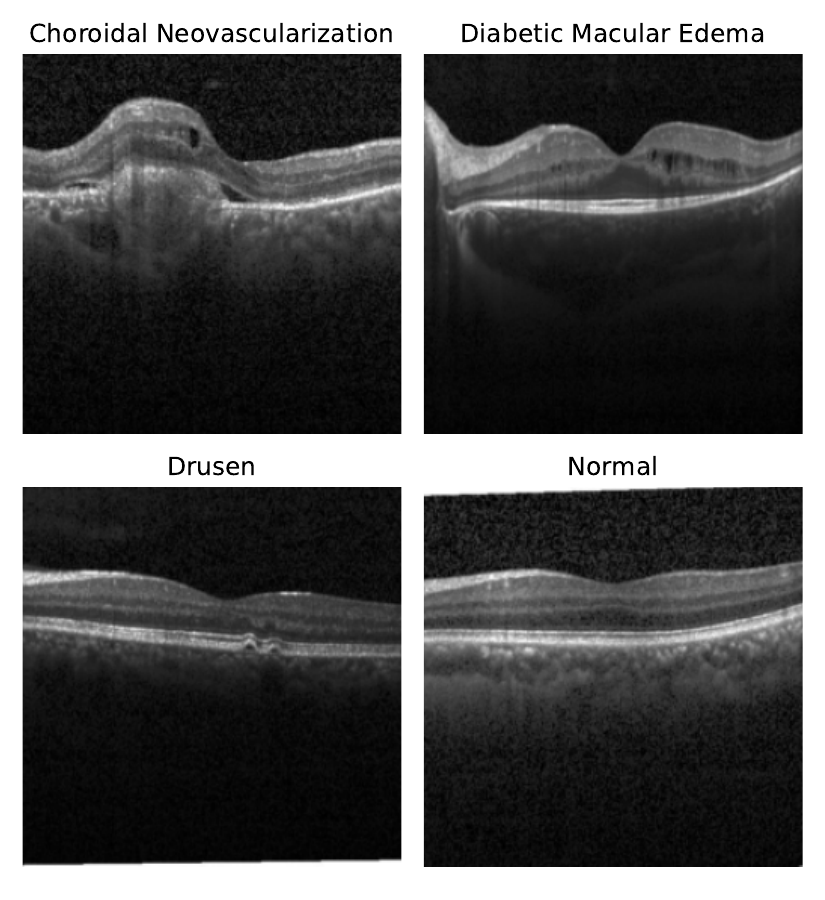}
\caption{Example of the images in the OCTMNISTv2 dataset.} 
\label{fig:data}
\end{figure}

We evaluated CoLoRA's fine-tuning performance in the task of image classification. We chose the problem of classifying optical coherence tomography (OCT) images of patients with Choroidal Neovascularization, Diabetic Macular Edema, Drusen, or Normal. Fig.~\ref{fig:data} presents images of the mentioned classes. Our experiments used the OCTMNISTv2 dataset \cite{medmnistv1,medmnistv2}. OCT imaging is now a standard tool for diagnosing and guiding the treatment of several major causes of blindness worldwide \cite{ferrara2010vascular}. OCTMNISTv2 is publicly available through MedMNIST and is highly imbalanced. For example, the Drusen class contains 7,754 training images, whereas the Normal class contains 46,026, corresponding to a ratio of approximately 6:1; see Table \ref{tab:octmnist}.

Hence, to favor the reproducibility of our experiment and keep the models as simple as possible, instead of resorting to training techniques with unbalanced classes (weighting of classes or data augmentation). To construct a balanced training set while preserving a simple and reproducible protocol, we selected 7,754 images from each class, corresponding to the size of the smallest class. Because the training samples follow the dataset's predefined ordering, this selection results in 31,016 training images from the original 97,477.

\begin{table}[ht]
    \centering
    \begin{tabular}{|cc|rcc|cc|}
    \hline
        Class    & Index & Train   & Val. & Test  & Bal. & Dst. \\
    \hline
        ChN      & 1     & 33,484  & 2708 & 250  & 7754 & 7040 \\
        DME      & 2     & 10,213  & 2708 & 250  & 7754 & 7040  \\
        Drusen   & 3     & 7754    & 2708 & 250  & 7754 & 7040  \\
        Normal   & 4     & 46,026  & 2708 & 250  & 7754 & 7040  \\
    \hline
    \end{tabular}
    \caption{Number of images in the subset OCTMNISTv2 Original dataset, and in the Balanced (Bal) and Distilled training sets; see text.}
    \label{tab:octmnist}
\end{table}

\subsection{Evaluation Metrics}
Given the confusion matrix $C\in\mathbb{R}^{4 \times 4}$ with $C_{i j}$ the number of data with true class $i$ predicted as $j$. We use the following one-vs-rest metrics for evaluating the methods' performance:

\begin{align}
    \text{Recall}_i    &= \frac{C_{i i}}{\sum_{j} C_{i j}}, \\
    \text{Precision}_i &= \frac{C_{i i}}{\sum_{j} C_{j i}}, \\
    \text{Accuracy}    &= \frac{Tr(C)}{\sum_{p, q} C_{p q}}, \\
    \text{Specificity}_i &= \frac{\sum_{p\neq i}\sum_{q\neq i} C_{p q}}{\sum_{p\neq i}\sum_{q} C_{p q}}, \\
    \text{F1}_i          &= \frac{2\,\text{Precision}_i\,\text{Recall}_i}{\text{Precision}_i+\text{Recall}_i};
\end{align}

\noindent where F1 indicates the F1-score. In addition we also use the ROC curves plot the $\text{TPR}_i / \text{Recall}_i$ vs. $\text{FPR}_i=1-\text{Specificity}_i$ as the decision threshold varies. 

\change{
    We conducted ten independent fine-tuning runs of the VGG16–CoLoRA model for a fixed schedule of 20 epochs. After each epoch, the CoLoRA weights were merged into the backbone, the depthwise kernel and bias were reset to zero, and the pointwise kernel was reinitialized. Validation accuracy was computed during training and used to retain the best checkpoint for each run. The held-out test set was not used for checkpoint selection, early stopping, hyperparameter optimization, or per-epoch monitoring. After the fixed training schedule was completed, the validation-selected checkpoint was restored and the held-out test set was evaluated once to obtain the reported final test metrics.
}
 

\subsection{VGG16-\CoLoRA{}}

\change{
    Following the protocol described above, ten independent VGG16–CoLoRA runs were trained for the complete 20-epoch schedule. The best checkpoint in each run was selected using validation accuracy, and the held-out test set was evaluated only after this checkpoint had been restored.
}

\begin{figure}[ht!]
    \centering
    \subfloat[Balanced dataset.]{%
        \includegraphics[width=0.48\textwidth]{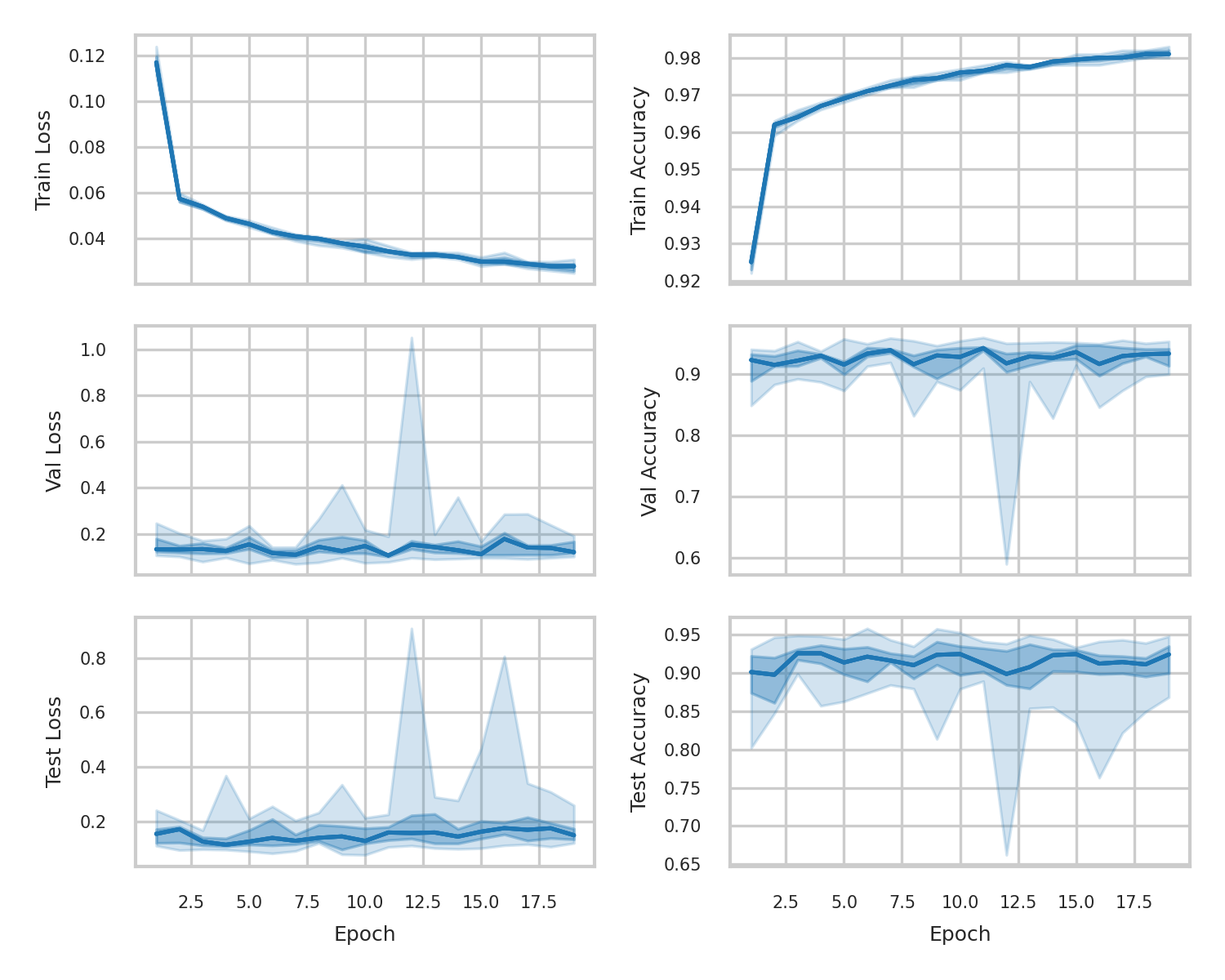}%
        \label{fig:vgg16_accuracies_balanced}}
    \\[1ex] 
    \subfloat[Distilled dataset.]{%
        \includegraphics[width=0.48\textwidth]{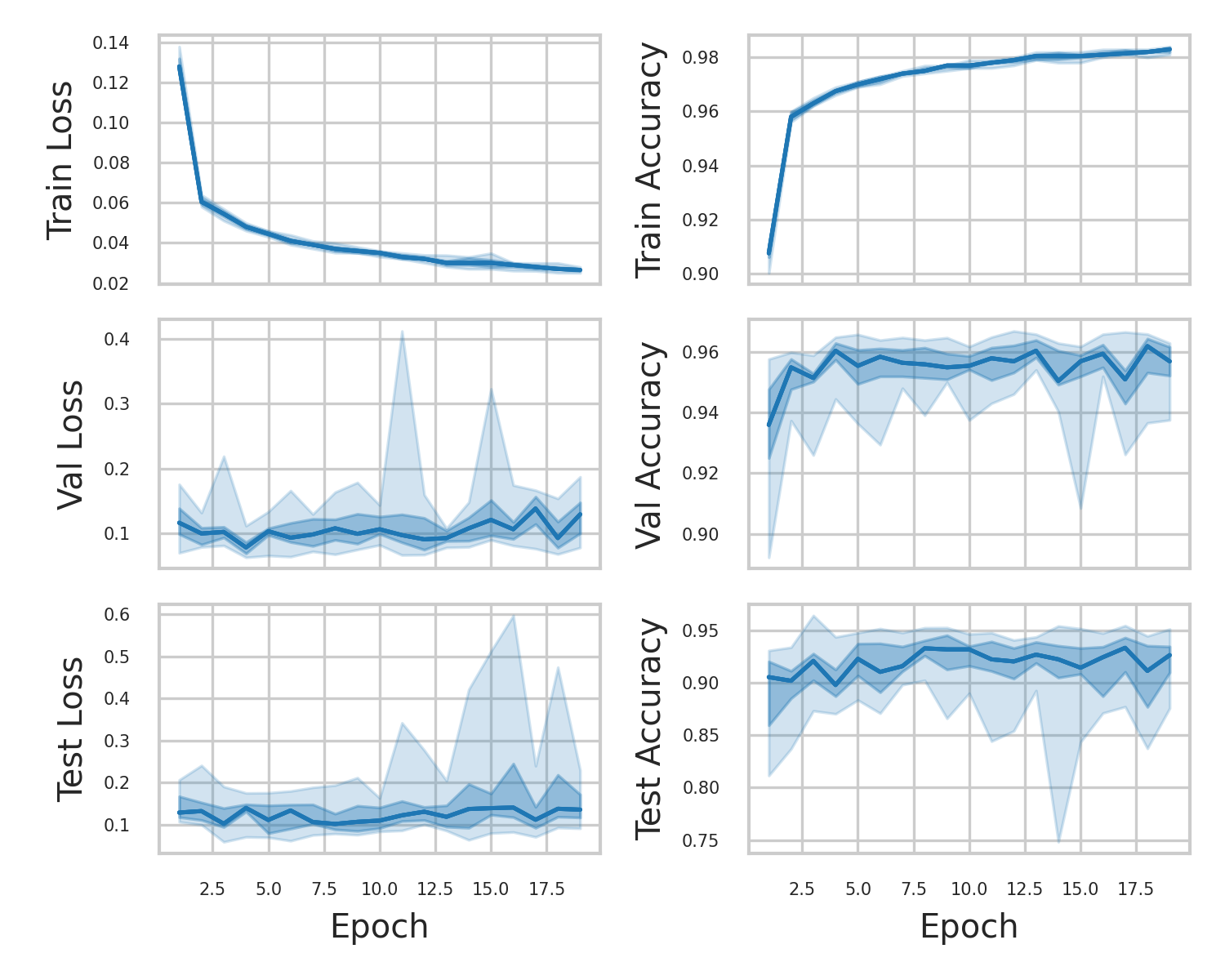}%
        \label{fig:vgg16_accuracies_distilled}}
    \caption{Median (solid line), interquartile range (IQR, dark band), and full range (light band) of the training and validation learning curves across the 10 independent VGG16–CoLoRA runs. Model selection is based on validation accuracy; the held-out test set is reserved for the final evaluation of the selected checkpoint.}
    \label{fig:vgg16_accuracies}
\end{figure}

Fig.~\ref{fig:vgg16_accuracies} shows median (solid line), interquartile range (IQR, dark band), and full range (light band) for the training and validation learning curves across the 10 independent runs. The relatively narrow IQR indicates stable behavior across most runs, although occasional runs exhibit substantially poorer performance at specific epochs. This variability highlights the importance of evaluating CoLoRA across multiple independent training runs rather than relying on a single realization.

Fig.~\ref{fig:vgg16_accuracies_balanced} shows the evolution of such metrics: the
bold line corresponds to the median value per epoch, the dark blue region allocates $50\%$ of the values closer to the median (interquartile range, IQR) and the light blue shows the full interval of the values. The loss and accuracy (ratio of correctly classified data) in the train dataset are smooth and consistent (similar to the experiments): the loss decreases, and the accuracy approaches one. This characterizes a correct parameter update by CoLoRA-based fine-tuning, the main contribution of this work. 

\begin{figure}[t!]
    \centering
    \subfloat[Balanced dataset.]{
            \includegraphics[width=0.48\textwidth]{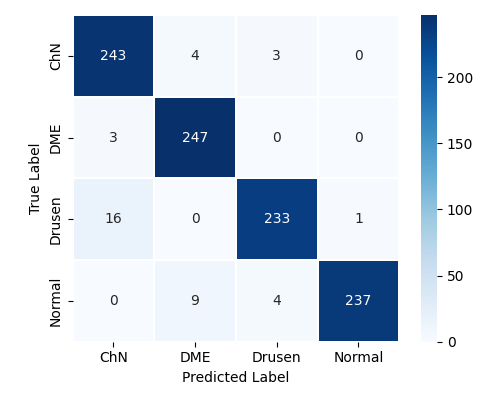}
            \label{fig:balanced_vgg16_confusion}
        }
    \hfill
    \\[1ex] 
    \subfloat[Distilled dataset.]{
            \includegraphics[width=0.48\textwidth]{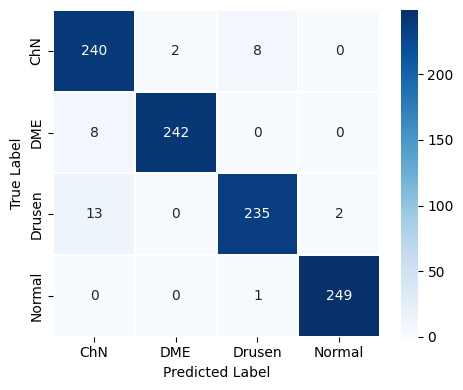}
            \label{fig:distilled_vgg16_confusion}
        }
    \caption{Confusion matrices on the OCTMNISTv2 test set for VGG16–CoLoRA.}
    \label{fig:vgg16_confusions}
\end{figure}

\begin{figure*}[!t]
    \centering
    \subfloat[{Balanced dataset}]{
            \includegraphics[width=0.9\linewidth]{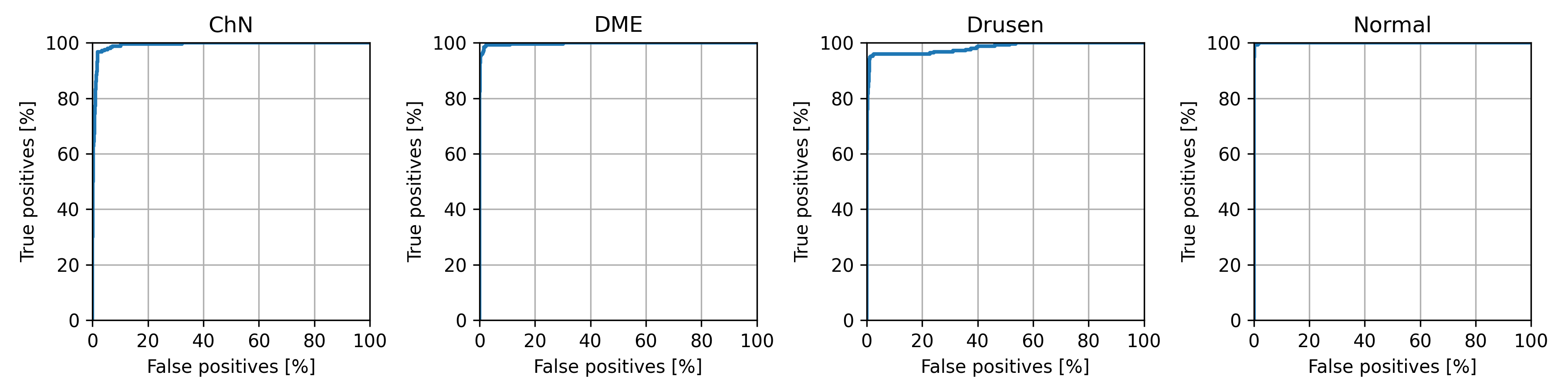}%
            \label{fig:balanced_roc_vgg16}
        }
    \hfill
    \subfloat[{Distilled dataset}]{
            \includegraphics[width=0.9\linewidth]{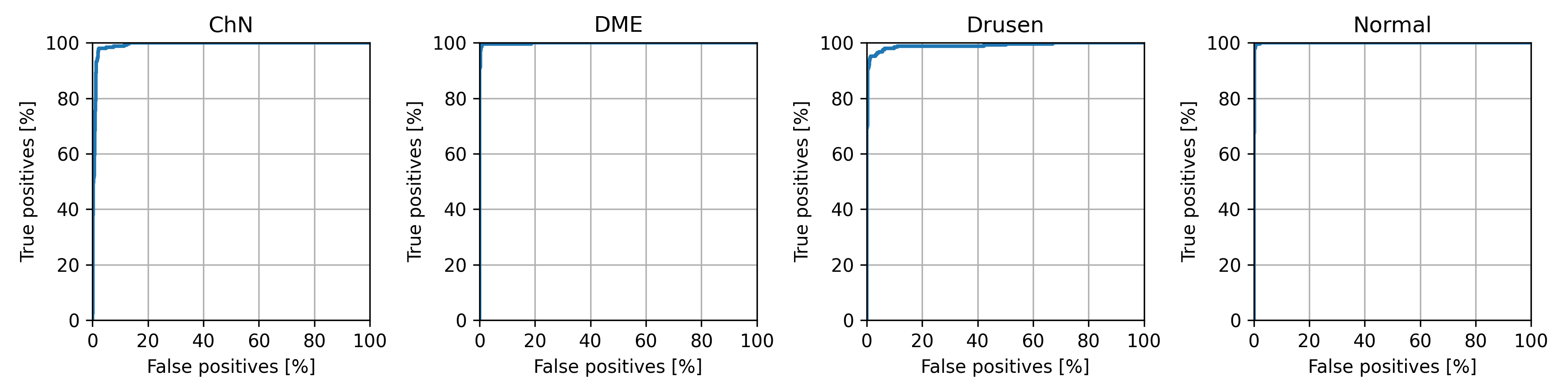}
            \label{fig:distilled_roc_vgg16}
        }
    \caption{One-vs-rest ROC curves for VGG16–CoLoRA on OCTMNISTv2.}
    \label{fig:vgg16_rocs}
\end{figure*}

To highlight the importance of having a computationally efficient and effective parameter tuning method as CoLoRA, it is essential to analyze the results on the selected task: the OCTMNISTv2 classification. We trained the model 10 times, randomly initializing it each time and randomly re-initializing the $K_p$ parameters at each epoch (after the parameter merging mentioned in section III.C). The training and validation loss/accuracy curves were monitored throughout the fixed schedule. \change{Across most runs, the validation metrics remain within a relatively narrow interquartile range, although a few runs exhibit substantially poorer performance at specific epochs.} There are epochs in which some experiments appear to be trapped at a local minimum with a clear overfit to the training data (lower-tail deviations/outlier runs). This underscores the importance of having a fast method for fine-tuning parameters and, therefore, being able to perform multiple training runs. \change{For each independent run, the checkpoints for each epoch were 
restored after training, and the held-out test set was then evaluated once to obtain the final reported performance.} The VGG16–CoLoRA confusion matrix is shown in Fig.~\ref{fig:balanced_vgg16_confusion}. Table \ref{tab:AUC_ROC_balanced_vgg16} presents the AUC-ROC and accuracy metrics. Then, Fig.~\ref{fig:balanced_roc_vgg16} shows the ROC curves.

\begin{table}[ht!]
    \centering
    \begin{tabular}{|c|cccc|}
    \hline
         Class     & AUC ROC & Recall & Precision & F1 \\
        \hline
        ChN      & 0.992 & 0.972 & 0.927 & 0.949 \\
        DME      & 0.998 & 0.988 & 0.950 & 0.969 \\
        Drusen   & 0.983 & 0.932 & 0.971 & 0.951 \\
        Normal   & 0.999 & 0.948 & 0.995 & 0.971 \\
        \hline
        Avg.     & 0.993 & 0.960 & 0.961 & 0.960 \\
        \hline
    \end{tabular}
    \caption{VGG16–CoLoRA: class-wise metrics on the balanced training regime (test set).}
    \label{tab:AUC_ROC_balanced_vgg16}
\end{table}

\begin{figure}[ht]
    \includegraphics[width=1\linewidth]{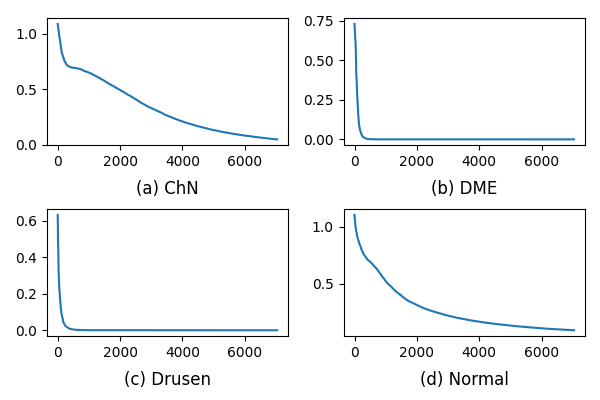}
    \caption{Sorted predictive entropy of the balanced training samples. The distilled subset retains 7,040 samples per class after removing samples from both extremes of the entropy distribution.}
    \label{fig:vgg16_uncertaint}
\end{figure}

\begin{figure}[t!]
    \centering
    \includegraphics[width=0.8\linewidth]{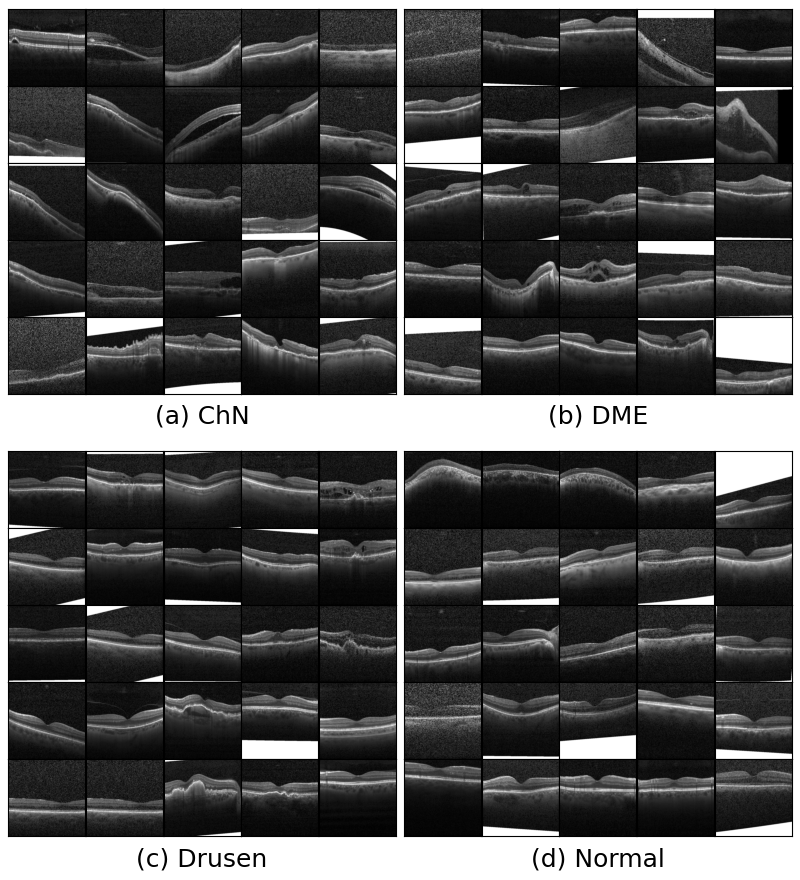}
    \caption{Highest-entropy samples.}
    \label{fig:entropy_bad}
\end{figure}

\begin{figure}[t!]
    \centering
    \includegraphics[width=0.8\linewidth]{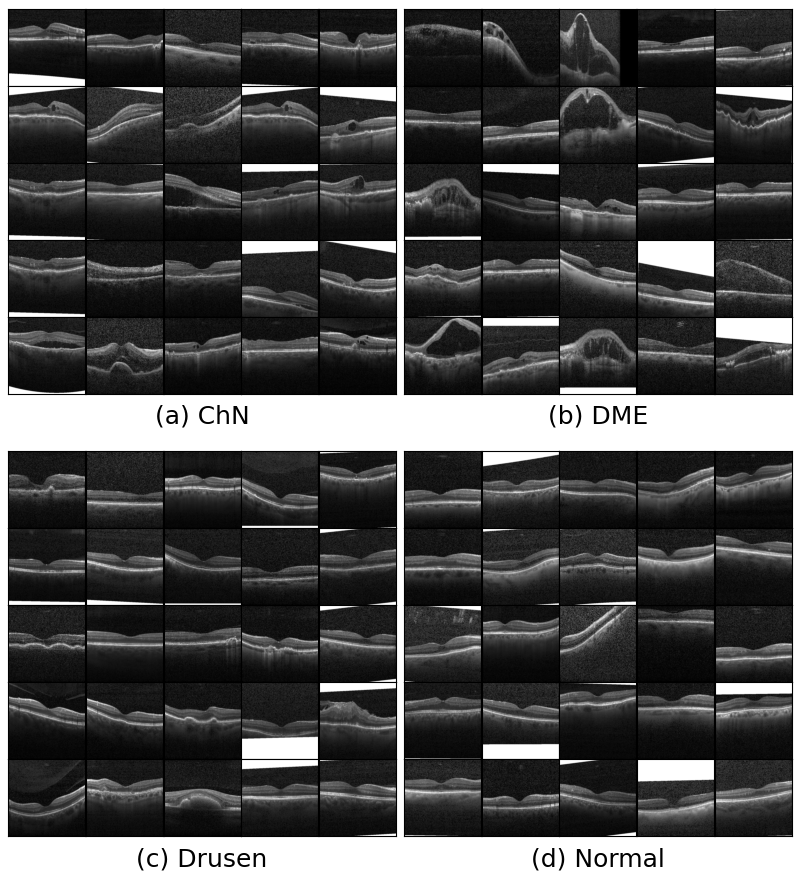}
    \caption{Low-entropy samples.}
    \label{fig:entropy_good}
\end{figure}

As mentioned, previously reported experiments are based on a balanced database using the first 7754 data points from each class (the number of cases in the smallest class). Hence, we employ a data distillation strategy to perform a more appropriate data selection. Therefore, we sort the training data by prediction entropy with the best pre-trained model.

\begin{figure}[t!]
\centering
    \includegraphics[width=0.5\linewidth]{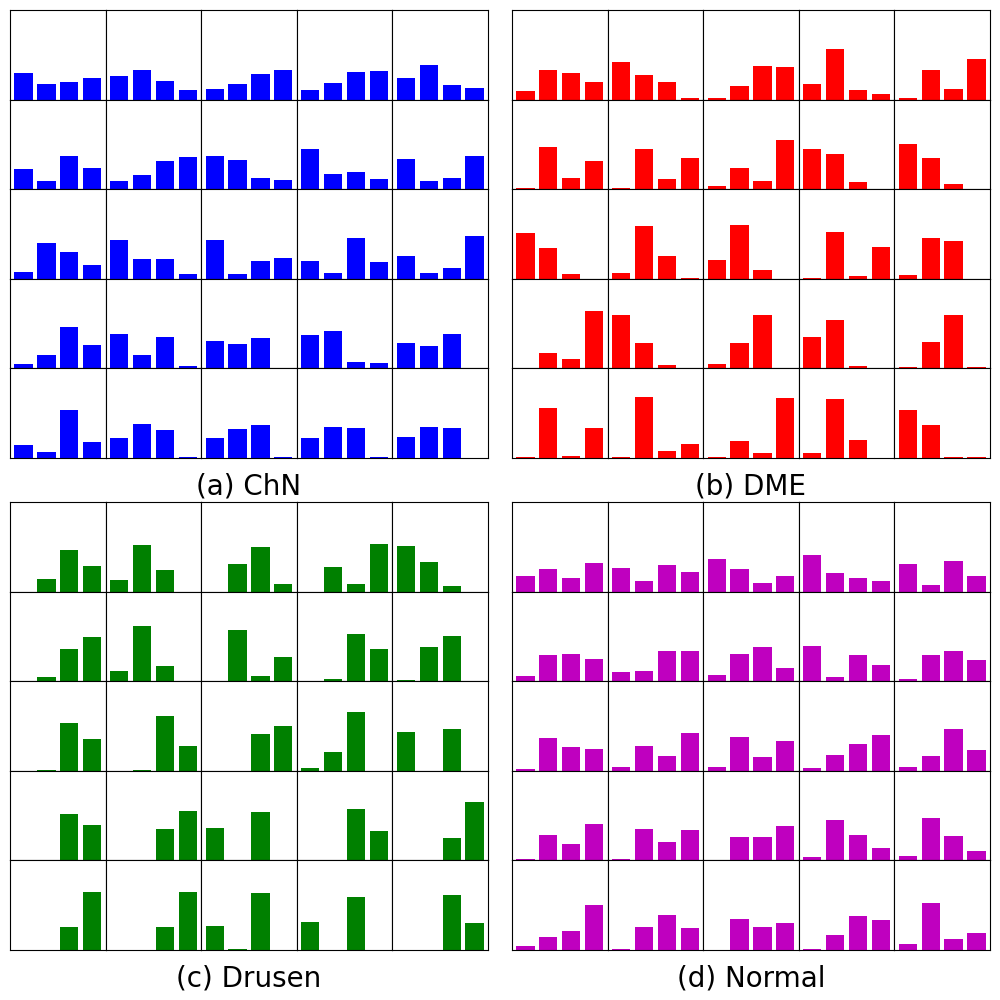}
    \caption{Predictions for highest-entropy samples in Fig.~\ref{fig:entropy_bad}.}
    \label{fig:predictions_bad}
\end{figure}

\begin{figure}
\centering
    \includegraphics[width=0.5\linewidth]{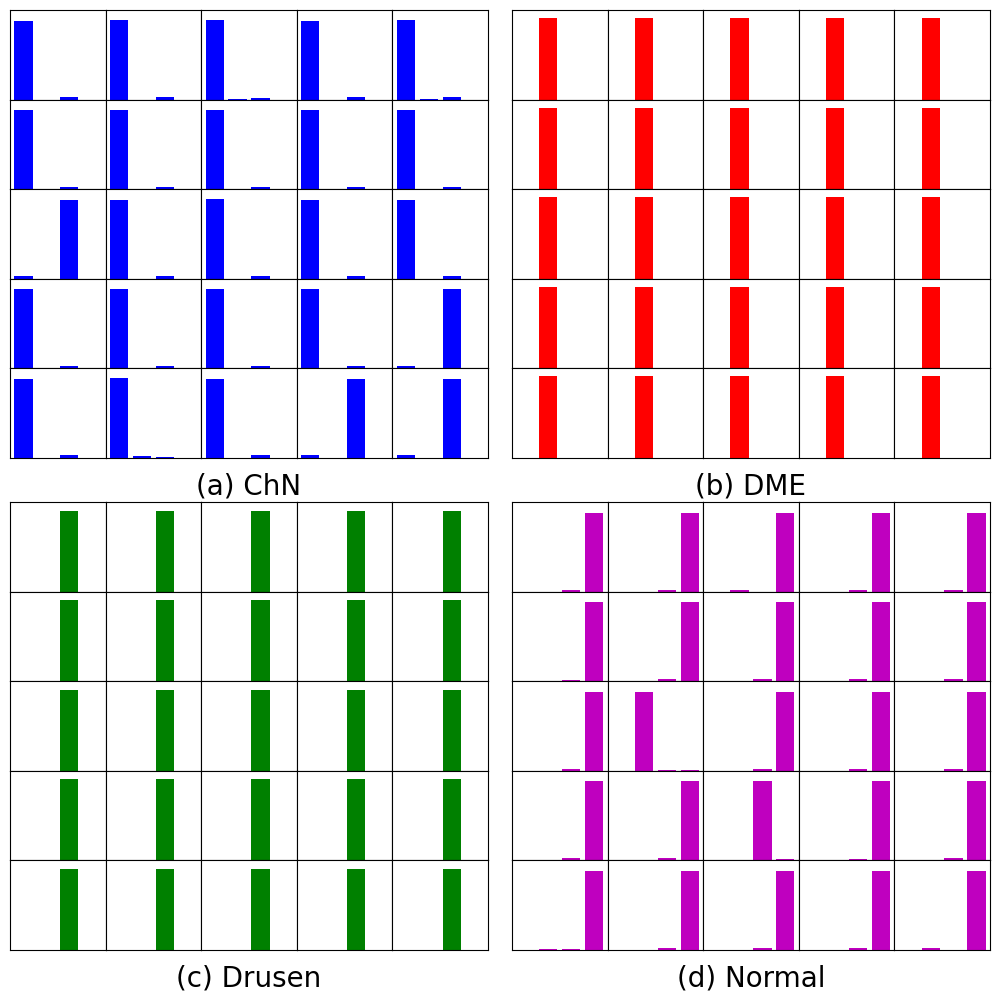}
    \caption{Predictions for low-entropy samples in Fig.~\ref{fig:entropy_good}.}
    \label{fig:predictions_good}
\end{figure}

Figures~\ref{fig:vgg16_uncertaint}--\ref{fig:predictions_good} are included to make this distillation step interpretable. Fig.~\ref{fig:vgg16_uncertaint} shows the sorted predicted entropy values. This reveals a small set of clearly uncertain samples with high-entropy region. Fig.~\ref{fig:entropy_bad} presents representative high-entropy images, while Fig.~\ref{fig:predictions_bad} depicts their corresponding model outputs. Together, these figures indicate that many of the most uncertain samples are also visually degraded or atypical, for example with noticeable rotation, displacement, or poor centering, which makes them plausible candidates for removal from the training set. In contrast, Fig.~\ref{fig:entropy_good} and Fig.~\ref{fig:predictions_good} show representative lower-entropy samples and their predictions, illustrating the kind of more stable and informative training examples retained after distillation. \change{Hence, these figures provide qualitative evidence that predictive entropy is associated with image difficulty and can be used as a practical criterion to refine the training subset.}

\change{Based on the predictive-entropy ranking, samples were sorted independently within each class in ascending order of entropy. From the 7,754 balanced training samples per class, we removed ranks 1--689 (the lowest-entropy samples) and ranks 7,730--7,754 (the 25 highest-entropy samples), retaining ranks 690--7,729. Thus, we removed the easiest samples and a small number of the most difficult samples, which were assumed to represent atypical cases that might not reflect the overall class distribution. The test dataset was not modified. The resulting distilled training dataset contained exactly 7,040 samples per class. This subset size, which is divisible by 128, was selected empirically rather than derived theoretically or optimized. It retained most of the available training data while yielding an exact number of mini-batches for the batch size of 8 used throughout the experiments. Consequently, the entropy-based distillation experiment is presented as an ancillary empirical analysis rather than as evidence that these thresholds are optimal. We then fine-tuned the CoLoRA model anew, keeping the backbone kernels frozen during each epoch and merging the learned updates after every epoch according to \eqref{eq:margeK0} and \eqref{eq:margeb0}. The results of this second fine-tuning stage were used as the final outcomes.}

\begin{table*}[ht!]
    \centering
    \begin{tabular}{|c|cccc|cccc|}
    \hline
         &  \multicolumn{4}{c|}{Transfer Learning}  &  \multicolumn{4}{c|}{CoLoRA Fine-Tuning}   \\
        \hline
        Class    & AUC & Rec. & Prec. & F1 & AUC & Rec. & Prec.  & F1\\
        \hline
        ChN     & 0.980 & 0.960 & 0.863 & 0.909 & 0.992 & 0.980 & 0.925 & 0.951\\
        DME     & 0.983 & 0.932 & 0.955 & 0.943 & 0.999 & 0.972 & 0.988 & 0.980\\
        Drusen  & 0.975 & 0.824 & 0.915 & 0.867 & 0.990 & 0.904 & 0.983 & 0.941\\
        Normal  & 0.993 & 0.948 & 0.937 & 0.942 & 0.998 & 0.992 & 0.958 & 0.975\\
        \hline
        Avg.    & 0.983 & 0.916 & 0.918 & 0.915 & \textbf{0.995} & \textbf{0.962} & \textbf{0.963} & \textbf{0.962}\\
        \hline
    \end{tabular}
    \caption{VGG16—Transfer vs.\ VGG16—CoLoRA on the distilled training set (test metrics).}
    \label{tab:AUC_ROC_distilled_vgg16}
\end{table*}

The final training is performed on the distilled data. \change{We performed ten independent CoLoRA runs for a fixed schedule of 20 epochs. For each run, the checkpoint achieving the highest validation accuracy was restored after training and the held-out test set was evaluated once.} The training and validation learning curves are shown in Fig.~\ref{fig:vgg16_accuracies_distilled}, the confusion matrix in Fig.~\ref{fig:distilled_vgg16_confusion} and the ROC curves in Fig.~\ref{fig:distilled_roc_vgg16}. Finally, Table \ref{tab:AUC_ROC_distilled_vgg16} presents the performance metrics of VGG16-CoLoRA for each class in the dataset. 

\change{
    Table \ref{tab:metric_vgg16} places our results in context with previously reported performance from recent state-of-the-art models, including MedViT (v1 \cite{manzari2023medvit} and v2 \cite{manzari2025medical}), MedKAN \cite{yang2025medkan}, and MedMamba \cite{yue2024medmamba}, based respectively on Vision Transformers, Kolmogorov-Arnold networks, and state-space models. These approaches are trained end-to-end and generally involve substantially more trainable parameters, whereas the VGG16-CoLoRA model contains 15.6 million total parameters, of which only 2.5 million are optimized by backpropagation while the remaining convolutional weights are updated through the merging procedure. Numerically, VGG16-CoLoRA is competitive with several of these reported results; however, because the literature models were not evaluated under the same experimental protocol, these comparisons are included only for contextual reference and should not be interpreted as direct head-to-head evaluations.
}

\subsection{\CoLoRA{} Layer Placement Study} 

\begin{figure}[ht]
\centering
    \includegraphics[width=0.7\linewidth]{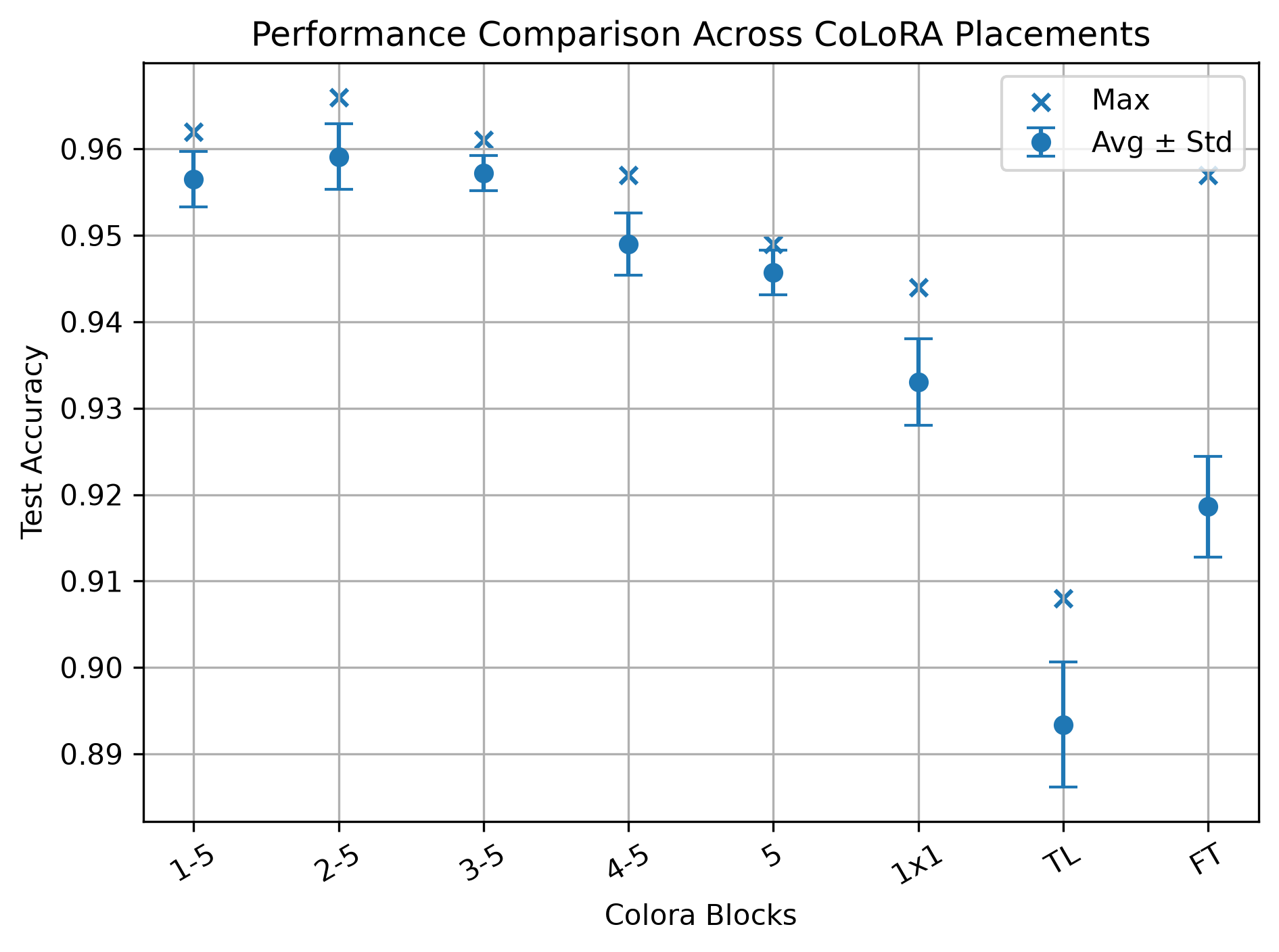}
    \caption{Mean test accuracy, standard deviation, and maximum test accuracy across the evaluated CoLoRA placement configurations and VGG16 baselines. The labels $1x1$, TL and FT corresponds to remove the dephwise kernel in the 1-5 configuration, perform an standard transfer learning procedure, and achive a full-tuning strategy; respectively.}
    \label{fig:vgg16_placement}
\end{figure}

To quantify the effect of CoLoRA layer placement, we evaluated five configurations in the VGG16 backbone: CoLoRA applied to convolutional blocks 1 to 5, 2 to 5, 3 to 5, 4 and 5, and only block 5. We additionally considered a simplified CoLoRA variant using only the pointwise kernel, as well as Transfer Learning and Full Fine-Tuning baselines. For the CoLoRA placement configurations, we used a common protocol with pretrained weights, no data augmentation, the same batch size, and validation-based checkpoint selection. The Transfer Learning and Full Fine-Tuning baselines used the same dataset, preprocessing, batch size, optimizer settings, and evaluation metrics. The placement configurations were predefined before evaluation, and test-set performance was not used to define the CoLoRA placement choices. The results are summarized numerically in Table~\ref{tab:placement_study} and visually in Fig.~\ref{fig:vgg16_placement}. All standard CoLoRA configurations improve substantially over Transfer Learning, increasing the average test accuracy from $0.894$ to values between $0.940$ and $0.959$. The highest average test accuracy is obtained when CoLoRA is applied to blocks 2 to 5, reaching $0.959 \pm 0.003$, together with the highest maximum test accuracy of $0.966$. Applying CoLoRA to blocks 3 to 5 produces a very similar average accuracy of $0.957 \pm 0.002$, while reducing the average epoch time from $3.00$ to $2.79$ minutes. 

\begin{table}[ht!]
    \centering
    \begin{tabular}{|l|c c c|}
        \hline
        Method   & AUC &  Acc. & Params.\\
        \hline
        ResNet-18  (28) \cite{medmnistv1}             & 0.951 & 0.758 & \\
        ResNet-50 (224) \cite{medmnistv1}             & 0.951 & 0.750 & \\
        Dedicated CNN (28) \cite{wilhelmi2024simple}  &       & 0.760 & \\
        ResNet-50 (224) \cite{wilhelmi2024simple}     &       & 0.776 & \\
        \hline
        MedViTv1-T  \cite{manzari2023medvit}          & 0.961 & 0.767 & 15.2M \\
        MedViTv1-S  \cite{manzari2023medvit}          & 0.960 & 0.782 & \\
        MedViTv1-L  \cite{manzari2023medvit}          & 0.945 & 0.761 & \\
        MedKAN-S \cite{yang2025medkan}                & 0.993 & 0.921  & 11.5M  \\
        MedKAN-B \cite{yang2025medkan}                & \textbf{0.996} & 0.927  & 24.6M \\
        MedKAN-L \cite{yang2025medkan}                & 0.994 & 0.925  & 48.0M \\
        MedMamba-T  \cite{yue2024medmamba}            & 0.992 & 0.918 & 15.2M \\
        MedMamba-S  \cite{yue2024medmamba}            & 0.991 & 0.929 & 23.5M\\
        MedMamba-B  \cite{yue2024medmamba}            & \textbf{0.996} & 0.927 & 48.1M\\
        MedMamba-X  \cite{yue2024medmamba}            & 0.993 & 0.928 & \\
        MedViTv2-T  \cite{manzari2025medical}         & 0.993 & 0.927 &  \\
        MedViTv2-S  \cite{manzari2025medical}         & 0.994 & 0.942 & \\
        MedViTv2-B  \cite{manzari2025medical}         & \textbf{0.996} & 0.944 & 32.3M\\
        MedViTv2-L  \cite{manzari2025medical}         & \textbf{0.996} & 0.952 & \\
        \hline
        ResNet50—Transfer Learning                     & 0.983 & 0.903 & 37.7M \\
        ResNet50—\textbf{CoLoRA}                       & 0.992 & 0.951 & 14.7M\textsuperscript{\dag} \\
        VGG16—Transfer Learning                        & 0.982 & 0.916 & 0.87M \\
        VGG16—\textbf{CoLoRA}                          & 0.995 & \textbf{0.966} & 2.54M\textsuperscript{\ddag} \\
        \hline
    \end{tabular}
    \caption{Comparison on OCTMNISTv2. \textsuperscript{\dag}\;14.7M parameters trained by backprop out of 37.7M total. \textsuperscript{\ddag}\;2.54M trained by backprop out of 15.6M total.}
    \label{tab:metric_vgg16}
\end{table}

The results reveal a clear efficiency--performance trade-off associated with layer placement. Applying CoLoRA to all convolutional blocks achieves a strong average test accuracy of $0.956 \pm 0.003$ and a maximum accuracy of $0.962$, but also incurs the highest computational cost among the CoLoRA configurations, requiring 2.55M trainable parameters and 4.42 minutes per epoch. In contrast, restricting adaptation to blocks 4 and 5 or only block 5 decreases the average epoch time to 2.15 and 1.64 minutes, respectively, at the expense of lower average accuracies of $0.940$ and $0.941$. Nevertheless, both configurations remain clearly above the Transfer Learning baseline. Interestingly, extending CoLoRA from blocks 2--5 to block 1 does not provide an additional performance benefit. The blocks 1--5 configuration obtains an average accuracy of $0.956$, compared with $0.959$ for blocks 2--5, despite requiring substantially more training time. This suggests that adapting the earliest convolutional block is not necessary for this task and that most of the useful adaptation can be achieved by modifying the intermediate and deeper feature representations. 

We also evaluated an ablation in which the depthwise CoLoRA kernel was removed and only the pointwise kernel of size $1 \times 1 \times C$ was retained. This simplified blocks 2--5 configuration requires only 0.91M trainable parameters and 1.95 minutes per epoch, but its average test accuracy decreases to $0.933 \pm 0.005$, with a maximum of $0.944$. Although this result remains above the Transfer Learning baseline, it is substantially below the complete blocks 2--5 CoLoRA configuration. 
\change{This comparison supports the contribution of the depthwise component, indicating that spatial adaptation is important for obtaining the full benefit of the proposed decomposition. Full Fine-Tuning achieves an average test accuracy of $0.918 \pm 0.005$ and a maximum of $0.957$, while updating the complete VGG16 model and requiring 3.27 minutes per epoch. The Full Fine-Tuning result is included as a direct VGG16 reference for the reviewer-requested comparison with unrestricted backbone adaptation. In comparison, the standard CoLoRA configurations achieve higher average accuracy while updating substantially fewer parameters. In particular, CoLoRA on blocks 2--5 increases the reported average accuracy from $0.918$ to $0.959$ while training only 2.54M parameters.}

Overall, Table~\ref{tab:placement_study} \change{and Fig.~\ref{fig:vgg16_placement}} show that the best predictive performance is obtained by adapting blocks 2--5, whereas configurations restricted to deeper blocks provide lower computational cost with a moderate reduction in accuracy. Therefore, the results suggest that CoLoRA should preferentially be applied to intermediate and deeper convolutional blocks, extending the adaptation toward earlier layers only when the validation improvement justifies the additional computational cost.

\begin{table}[t]
    \centering
    \scriptsize
    \setlength{\tabcolsep}{4pt}
    \renewcommand{\arraystretch}{1.05}
    \begin{tabular}{|l|cccc|}
        \hline
        CoLoRA    & Trainable & Avg. epoch & Test Acc.     & Max. Test  \\
        Placement & Params.  & Time (min) & Mean $\pm$ SD & Acc. \\
        \hline
        1 to 5              & 2.55M & 4.42 & $0.956 \pm 0.003$ & 0.962 \\
        2 to 5              & 2.54M & 3.00 & $\bf{0.959} \pm 0.003$ & \bf{0.966} \\
        3 to 5              & 2.51M & 2.79 & $0.957 \pm 0.002$ & 0.961 \\
        4 to 5              & 2.34M & 2.15 & $0.940 \pm 0.003$ & 0.957 \\
        Only 5              & 1.67M & 1.64 & $0.941 \pm 0.002$ & 0.949 \\
        \hline
        2 to 5 $(1 \times 1)$ & 0.91M  & 1.95 & $0.933 \pm 0.005$ & 0.944 \\
        Transfer Learning     & 0.87M  & 1.32 & $0.894 \pm 0.007$ & 0.908 \\
        Full Fine-Tuning           & 15.58M & 3.27 & $0.918 \pm 0.005$ & 0.957 \\
        \hline
    \end{tabular}
\caption{Effect of CoLoRA layer placement on VGG16 performance and training efficiency. Test accuracy is reported as mean $\pm$ standard deviation together with the maximum observed accuracy. The pointwise-only CoLoRA variant, Transfer Learning, and Full Fine-Tuning are included as reference baselines.}
\label{tab:placement_study}
\end{table}

\subsection{Other MedMNIST Datasets}
We evaluated CoLoRA on additional MedMNIST datasets with the goal of not only reporting absolute results, but also situating VGG16--CoLoRA relative to recent methods founded on substantially different design principles. The comparison includes large end-to-end architectures such as MedViTv2 and MedMamba, whose reported variants range from 15.2M to 48.1M trainable parameters, as well as the quantum-hardware-based IBM Cleveland approach \cite{Singh2026MedMNIST}, which uses only 60 parameters on RetinaMNIST and 120 on PathMNIST and BloodMNIST, albeit with limited accuracy. It also includes MTDL \cite{Liu2026MTDL}, a specialized geometric-topological method that attains the best results on RetinaMNIST and strong results on PathMNIST with only 0.5M trainable parameters. In contrast, VGG16--Transfer Learning and VGG16--CoLoRA require only 0.9M and 2.6M trainable parameters, respectively. According to our results summarized in Table \ref{tab:metric_vgg16_other_db}, within this setting, CoLoRA consistently improves upon conventional transfer learning across all three datasets, raising the AUC/Acc. from 0.756/0.527 to 0.776/0.552 on RetinaMNIST, from 0.981/0.876 to 0.993/0.956 on PathMNIST, and from 0.995/0.932 to 0.998/0.981 on BloodMNIST. Although CoLoRA does not surpass the strongest transformer- or state-space-based models in these benchmarks, it remains highly competitive while relying on far fewer trainable parameters, and it clearly exceeds the quantum baseline in all cases. Beyond the numerical comparison, CoLoRA offers an important practical advantage: it adapts a pretrained CNN by updating only a small subset of parameters and then merges those updates into the original convolutional weights, thereby preserving the inference graph and deployment cost of the base model.

These literature-reported results are included only for contextual reference and are not part of the controlled PEFT comparison presented in Sec. IV-F.

\begin{table}[ht!]
    \centering
    \begin{tabular}{|l|c c c|}
        \hline
        \textbf{RetinaMNIST}  & AUC &  Acc. & Params.\\
        \hline
        Baseline                               & 0.750 & 0.531 & \\
        MedMamba-T  \cite{yue2024medmamba}     & 0.741 & 0.543 & 15.2M \\
        MedMamba-S  \cite{yue2024medmamba}     & 0.718 & 0.545 & 23.5M\\
        MedMamba-B  \cite{yue2024medmamba}     & 0.715 & 0.553 & 48.1M\\
        MedMamba-X  \cite{yue2024medmamba}     & 0.719 & 0.570 & \\
        MedViTv2-T  \cite{manzari2025medical}  & 0.761 & 0.547 & \\
        MedViTv2-S  \cite{manzari2025medical}  & 0.780 & 0.562 & \\
        MedViTv2-B  \cite{manzari2025medical}  & 0.783 & 0.575 & \\
        MedViTv2-L  \cite{manzari2025medical}  & 0.785 & 0.578 & 32.3M\\
        IBM Cleveland \cite{Singh2026MedMNIST} & 0.655 & 0.517 & 60   \\
        MTDL \cite{Liu2026MTDL}                & \textbf{0.874} & \textbf{0.665} & 0.5M \\
        VGG16—Transfer Learning                & 0.756 & 0.527 & 0.9M \\
        VGG16—CoLoRA                           & 0.776 & 0.552 & 2.6M\textsuperscript{\ddag} \\
        \hline
        \textbf{PathMNIST}  &  &  & \\
        \hline
        Baseline                               & 0.990 & 0.911 & \\
        MedMamba-T  \cite{yue2024medmamba}     & 0.997 & 0.953 & 15.2M \\
        MedMamba-S  \cite{yue2024medmamba}     & 0.997 & 0.955 & 23.5M\\
        MedMamba-B  \cite{yue2024medmamba}     & \textbf{0.999} & 0.964 & 48.1M\\
        MedMamba-X  \cite{yue2024medmamba}     & \textbf{0.999} & 0.962 & \\
        MedViTv2-T  \cite{manzari2025medical}  & 0.998 & 0.959 &  \\
        MedViTv2-S  \cite{manzari2025medical}  & 0.998 & 0.965 & \\
        MedViTv2-B  \cite{manzari2025medical}  & \textbf{0.999} & 0.970 & 32.3M\\
        MedViTv2-L  \cite{manzari2025medical}  & \textbf{0.999} & \textbf{0.977} & \\
        IBM Cleveland \cite{Singh2026MedMNIST} & 0.784 & 0.372 & 120  \\
        MTDL \cite{Liu2026MTDL}                & 0.996 & 0.920 & 0.5M \\
        VGG16—Transfer Learning                & 0.981 & 0.876 & 0.9M \\
        VGG16—CoLoRA                           & 0.993 & 0.956 & 2.6M\textsuperscript{\ddag} \\
        \hline
        \textbf{BloodMNIST}  &  &  & \\
        \hline
        Baseline                               & 0.998 & 0.966  & \\ 
        MedMamba-T  \cite{yue2024medmamba}     & \textbf{0.999} & 0.978 & 15.2M \\
        MedMamba-S  \cite{yue2024medmamba}     & \textbf{0.999} & 0.984 & 23.5M\\
        MedMamba-B  \cite{yue2024medmamba}     & \textbf{0.999} & 0.983 & 48.1M\\
        MedMamba-X  \cite{yue2024medmamba}     & \textbf{0.999} & 0.984 & \\
        MedViTv2-T  \cite{manzari2025medical}  & \textbf{0.999} & 0.979 & \\
        MedViTv2-S  \cite{manzari2025medical}  & \textbf{0.999} & 0.985 & \\
        MedViTv2-B  \cite{manzari2025medical}  & \textbf{0.999} & 0.985 & 32.3M\\
        MedViTv2-L  \cite{manzari2025medical}  & \textbf{0.999} & \textbf{0.987} & \\
        IBM Cleveland \cite{Singh2026MedMNIST} & 0.821 & 0.482 & 120  \\
        VGG16—Transfer Learning                & 0.995 & 0.932 & 0.87M \\
        VGG16—CoLoRA                           & 0.998 & 0.981 & 2.54M\textsuperscript{\ddag} \\
        \hline
    \end{tabular}
    \caption{Comparison on RetinaMNIST, PathMNIST, and BloodMNIST datasets. \textsuperscript{\ddag}\;2.6M trained by backprop out of 15.6M total.}
    \label{tab:metric_vgg16_other_db}
\end{table}

\subsection{\CoLoRA{} against other PEFT}

Table~\ref{tab:peft_vgg16} compares \CoLoRA{} with Transfer Learning and three CNN-compatible PEFT baselines in a VGG16 architecture: Adapters, BitFit~\cite{bitfit}, and Conv-LoRA~\cite{convlora}. The balanced OCTMNISTv2 setting provides the controlled comparison: all methods use the same data split, preprocessing, batch size, optimizer settings, evaluation definitions, and matched sequence of run seeds. The distilled setting is reported separately for Transfer Learning and CoLoRA as an additional analysis and is therefore not used for direct comparison across all PEFT methods.

\change{
    To ensure a common experimental framework, the comparison methods were reimplemented in TensorFlow/Keras. BitFit and Conv-LoRA were adapted from their original implementations and methodological descriptions, whereas the CNN adapter followed the formulation of Rebuffi et al. These adaptations preserve the defining update mechanism of each method while allowing all baselines and CoLoRA to be trained and evaluated using the same software framework. The exact implementations used in our experiments are available in the accompanying repository.
}

On balanced OCTMNISTv2, Adapters obtained the strongest predictive performance, reaching the highest AUC of 0.995 and the highest accuracy of 0.957. However, this required 1.50M trainable parameters and 1.56 minutes per epoch. BitFit used only 0.87M trainable parameters, matching the parameter count of standard Transfer Learning, but its lower accuracy of 0.896 suggests that bias-only adaptation was insufficient for this task. Conv-LoRA achieved a better trade-off than BitFit, reaching up to 0.949 accuracy and 0.993 AUC with fewer than 1M trainable parameters, although it required longer training times and depended on the selection of rank and scaling factor.

In the balanced setting, \CoLoRA{} did not achieve the highest overall accuracy, but the block 2--5 configuration outperformed all Conv-LoRA variants, reaching 0.954 accuracy and 0.994 AUC. This improvement came with a higher training cost of 3.00 minutes per epoch and 2.54M trainable parameters. Therefore, \CoLoRA{} is not the cheapest method during training; instead, its main advantage is its convolution-aware formulation and deployment simplicity. Unlike Conv-LoRA, \CoLoRA{} is defined directly from the original convolutional operator, without rank or scaling-factor hyperparameters, and its update can be merged into the pretrained backbone after training, preserving the original inference graph.

For fairness, the main comparison was conducted using the balanced dataset, since this setting allows all PEFT methods to be evaluated under the same data distribution and training protocol. Although the best overall result obtained with \CoLoRA{} was achieved on the distilled version of the dataset, this setting was not extended to every PEFT baseline because each method would require its own distilled-data generation and additional hyperparameter-specific training runs. In particular, methods such as Conv-LoRA would require separate evaluations for different rank and scaling-factor configurations, making a complete distilled comparison prohibitively time-consuming.

On the distilled dataset, \CoLoRA{} improved substantially over its balanced-dataset counterpart. In particular, \CoLoRA{} block 2--5 achieved the best accuracy in the table, reaching 0.966 accuracy and 0.994 AUC, while \CoLoRA{} block 5 obtained 0.959 accuracy and 0.993 AUC. These results suggest that \CoLoRA{} can benefit from improved data quality. Overall, Adapters remain the strongest baseline in raw balanced-dataset performance, while \CoLoRA{} provides a competitive alternative to Conv-LoRA by achieving comparable or improved predictive performance in the evaluated settings, avoiding an explicit per-update rank hyperparameter, and preserving a mergeable convolutional structure.

\begin{table}[ht!]
    \centering
    \setlength{\tabcolsep}{2.5pt} 
    \renewcommand{\arraystretch}{1.05} 
    \begin{tabular}{|c|l|c c c c|}
    \hline
    & Method & 
    \begin{tabular}{c}Avg epoch\\Time Mins\end{tabular} &
    \begin{tabular}{c}Test\\AUC\end{tabular} &
    \begin{tabular}{c}Test\\Acc.\end{tabular} &
    \begin{tabular}{c}Trained\\Params.\end{tabular} \\
    \hline
    & Baseline & & 0.958 & 0.776 & \\
    \hline
    \multirow{8}{*}{\rotatebox[origin=c]{90}{Balanced dataset}}
    & Transfer Learning & 1.32 & 0.979 & 0.933 & 0.87M \\
    & Adapters & 1.56 & \textbf{0.995} & \textbf{0.957} & 1.5M \\
    & BitFit & 1.71 & 0.979 & 0.896 & 0.87M \\
    & Conv-LoRA R4 $\alpha 8$ & 2.10 & 0.991 & 0.944 & 0.90M \\
    & Conv-LoRA R8 $\alpha 16$ & 2.25 & 0.990 & 0.943 & 0.93M \\
    & Conv-LoRA R16 $\alpha 32$ & 2.35 & 0.993 & 0.949 & 0.99M \\
    & CoLoRA block 5 & 1.64 & 0.993 & 0.939 & 1.67M$^\ddagger$ \\
    & CoLoRA block 2-5 & 3.00 & 0.994 & 0.954 & 2.54M$^\ddagger$ \\
    \hline
    \multirow{3}{*}{\rotatebox[origin=c]{90}{Distilled}}
    & Transfer Learning & 1.32 & 0.982 & 0.932 & 0.87M \\
    & CoLoRA block 5 & 1.64 & 0.993 & 0.959 & 1.67M$^\ddagger$ \\
    & CoLoRA block 2-5 & 3.00 & \textbf{0.994} & \textbf{0.966} & 2.54M$^\ddagger$ \\
    \hline
    \end{tabular}
    \caption{Performance comparison between PEFT methods and \CoLoRA{} in a VGG16 architecture on balanced and distilled OCTMNISTv2.}
    \label{tab:peft_vgg16}
\end{table}

To complement the comparison based on trainable parameters and training time, we measured the actual peak GPU memory consumption during backpropagation. All methods were evaluated under the same conditions using an NVIDIA GeForce RTX 3090, an input resolution of $224\times224$, a batch size of 8, float32 precision, 20 warm-up steps, and 200 measured training steps. Gradient checkpointing was not employed. Full fine-tuning, in which all 15.58M model parameters were updated, required a peak memory of 3.320~GiB. Head-only transfer learning presented the lowest peak consumption at 2.316~GiB, corresponding to a 30.2\% reduction relative to full fine-tuning. Adapters and BitFit required 2.518 and 2.670~GiB, representing reductions of 24.2\% and 19.6\%, respectively, whereas Conv-LoRA with $r=8$ and $\alpha=16$ required 3.052~GiB, an 8.1\% reduction. For CoLoRA, memory consumption depended strongly on the placement of the adaptation layers. Applying CoLoRA to blocks 2--5, the configuration that achieved the best predictive performance in the placement study, required 3.172~GiB, corresponding to a 4.4\% reduction relative to full fine-tuning. Applying CoLoRA to blocks 3--5, blocks 4--5, or only block 5 required 2.989, 2.951, and 3.001~GiB, respectively. In contrast, applying CoLoRA to all convolutional blocks required 3.639~GiB, exceeding full fine-tuning by 9.6\%. Notably, the best-performing CoLoRA blocks 2--5 configuration did not achieve the lowest memory footprint among the evaluated PEFT methods: Conv-LoRA required 3.052 GiB compared with 3.172 GiB for CoLoRA blocks 2--5. This further illustrates that the main advantage of CoLoRA lies in its structured parameter-efficient and mergeable adaptation rather than in minimizing peak training memory.

These measurements confirm that reducing the number of trainable parameters does not necessarily produce a proportional reduction in peak training memory. Activation retention, temporary convolutional outputs, optimizer states, and the spatial resolution of the adapted feature maps also contribute substantially to the runtime footprint. In particular, adapting earlier convolutional blocks increases memory consumption because these layers operate on higher-resolution feature maps. Consequently, CoLoRA should primarily be characterized as a parameter-efficient and deployment-efficient method, while its training-memory requirements depend on layer placement. These results also reinforce the practical criterion derived from the placement study: adapting deeper convolutional blocks generally provides a more favorable balance among predictive performance, trainable parameters, training time, and peak GPU memory.

\subsection{ResNet50-\CoLoRA{}}

Under matched settings to VGG16, we first froze the ResNet50 backbone and trained only the head; this underperformed relative to \cite{medmnistv1}. Full fine-tuning improved accuracy to 0.903 but required training 37M parameters. With CoLoRA and a frozen backbone, selective CoLoRA layers achieved 0.951 accuracy, substantially improving efficiency and performance. While strong, this did not surpass VGG16--CoLoRA's best accuracy. For residual architectures, a more extensive study is still needed to determine the most appropriate CoLoRA placement strategy. In particular, it remains to be analyzed whether CoLoRA should be applied uniformly across residual blocks, restricted to specific stages, or avoided in residual shortcut connections, since the optimal placement may differ from the behavior observed in VGG16. Finally, Table~\ref{tab:metric_vgg16} compares the investigated CoLoRA variants with baselines and SOTA models.

\subsection{Evaluation on Non-Medical Image Datasets}

We conducted complementary experiments on two non--medical image--classification benchmarks, CIFAR-100 and ImageNet--R, to evaluate the applicability of CoLoRA beyond the medical--imaging domain. These experiments are intended as a cross-domain validation rather than an exhaustive optimization of these benchmarks.

On CIFAR-100, all methods were trained for 100 epochs, and the results are summarized in Table~\ref{tab:metrics_cifar100}. The reported values correspond to the mean and standard deviation across the independent runs. Among the
evaluated methods, CoLoRA applied to blocks 2--5 achieved the highest mean performance, with an AUC of $0.961 \pm 0.000$, a Top--1 accuracy of $0.451 \pm 0.004$, and a Top--5 accuracy of $0.766 \pm 0.003$. Adapters and Conv-LoRA also obtained competitive results, reaching Top--1 accuracies of $0.449 \pm 0.004$ and $0.424 \pm 0.006$, respectively. In contrast, restricting
CoLoRA adaptation to block 5 resulted in an AUC of $0.911 \pm 0.004$, a Top--1 accuracy of $0.324 \pm 0.003$, and a Top--5 accuracy of $0.621 \pm 0.009$. These results indicate that, in this setting, extending CoLoRA adaptation across blocks 2--5 provides a substantial advantage over restricting the adaptation to the final convolutional block.

For a more challenging multiclass evaluation, we experimented with ImageNet--R. This dataset contains renditions of 200 ImageNet classes, including paintings, cartoons, and sketches, and was introduced to evaluate robustness under substantial changes in visual representation \cite{Hendrycks2021ManyFaces}. It has also been adopted in subsequent adaptation and continual-learning studies \cite{Wang2022DualPrompt,Liang2024InfLoRA}. Because ImageNet--R is class imbalanced and does not provide predefined training, validation, and test partitions, we constructed a class-balanced subset using 35 images per class for training, 5 for validation, and 10 for testing, resulting in 7,000 training, 1,000 validation, and 2,000 test images. Each method was trained in
10 independent runs for 50 epochs using the same class-balanced split and the same sequence of run seeds across methods. Table~\ref{tab:metrics_imagenet_r} reports the mean and standard deviation of the resulting test metrics across runs. We use these repeated-run statistics descriptively to characterize run-to-run variability and do not interpret the observed differences as formal hypothesis-test significance.

CoLoRA applied to blocks 2--5 achieved the highest mean performance among the evaluated methods, with an AUC of $0.754 \pm 0.008$, a Top--1 accuracy of $0.109 \pm 0.005$, and a Top--5 accuracy of $0.241 \pm 0.012$. Transfer Learning obtained $0.737 \pm 0.009$, $0.087 \pm 0.008$, and $0.210 \pm 0.009$, respectively, whereas Conv-LoRA obtained $0.697 \pm 0.016$, $0.071 \pm 0.009$, and $0.174 \pm 0.014$. These repeated-run results show that CoLoRA obtained the highest mean values among the three evaluated methods under this controlled ImageNet--R setting. The earlier anomalous single-run Conv-LoRA result is therefore not used as evidence in the revised analysis; the discussion is based on the repeated-run summaries shown in Table~\ref{tab:metrics_imagenet_r}.

The relatively modest absolute accuracies reflect the difficulty of ImageNet--R and the intentionally limited training subset of 35 images per class. Larger training subsets, longer schedules, stronger data augmentation, or additional hyperparameter optimization could potentially improve these results. However, such optimization is outside the scope of this work; our
objective is to evaluate whether CoLoRA remains effective when transferred to a challenging non-medical multiclass domain. Together with the CIFAR-100 experiments, these results provide additional evidence that CoLoRA can be applied beyond medical-image classification, while also showing that its performance depends on the selected adaptation layers and target domain.

\begin{table}[ht!]
    \centering
    \scriptsize
    \setlength{\tabcolsep}{2pt}
    \renewcommand{\arraystretch}{1.05}
    \begin{tabular}{|l|c c c c|}
        \hline
        Method & AUC & Top-1 & Top-5 & Params.\\
        \hline
        Transfer Learning & $0.889 \pm 0.004$ & $0.258 \pm 0.002$ & $0.535 \pm 0.004$ & 0.09M \\
        Adapters          & $0.942 \pm 0.004$ & $0.449 \pm 0.004$ & $0.730 \pm 0.008$ & 0.74M \\
        BitFit            & $0.914 \pm 0.003$ & $0.327 \pm 0.003$ & $0.613 \pm 0.003$ & 0.09M \\
        Conv-LoRA         & $0.947 \pm 0.002$ & $0.424 \pm 0.006$ & $0.732 \pm 0.008$ & 0.15M \\
        CoLoRA block 2-5  & $\bf{0.961} \pm 0.001$ & $\bf{0.451} \pm 0.004$ & $\bf{0.766} \pm 0.003$ & 1.77M \\
        CoLoRA block 5    & $0.911 \pm 0.004$ & $0.324 \pm 0.003$ & $0.621 \pm 0.009$ & 0.89M \\
        \hline
    \end{tabular}
    \caption{Performance comparison on CIFAR-100. Values are reported as mean $\pm$ standard deviation across the independent runs.}
    \label{tab:metrics_cifar100}
\end{table}

\begin{table}[ht!]
    \centering
    \scriptsize
    \setlength{\tabcolsep}{2pt}
    \renewcommand{\arraystretch}{1.05}

    \begin{tabular}{@{}|l|cccc|@{}}
        \hline
        Method & AUC & Top-1 & Top-5 & Params. \\
        \hline
        Transfer Learning & $0.737 \pm 0.009$ & $0.087 \pm 0.008$ & $0.210 \pm 0.009$ & 0.90M \\
        Conv-LoRA         & $0.697 \pm 0.016$ & $0.071 \pm 0.009$ & $0.174 \pm 0.014$ & 0.95M \\
        CoLoRA block 2-5  & $\bf{0.754} \pm 0.008$ & $\bf{0.109} \pm 0.005$ & $\bf{0.241} \pm 0.012$ & 2.57M \\
        \hline
    \end{tabular}

    \caption{Performance comparison on the class-balanced ImageNet-R subset using 35 training, 5 validation, and 10 test images per class. Values are reported as mean $\pm$ standard deviation across 10 independent runs.}
    \label{tab:metrics_imagenet_r}
\end{table}

\section{Conclusions}
\label{sec:conclusions}

We presented CoLoRA, a parameter-efficient fine-tuning strategy that extends LoRA to CNNs by decomposing convolutional updates into depthwise and pointwise components. CoLoRA can reduce the number of trainable convolutional-update parameters by more than 80\% relative to full convolutional fine-tuning and preserves the original inference structure through kernel-and-bias merging. On OCTMNISTv2, VGG16–CoLoRA achieved a best accuracy of 0.966 and an AUC of 0.995, while ResNet50–CoLoRA reached an accuracy of 0.951 with substantially fewer trainable parameters than full fine-tuning. Additional MedMNIST experiments further showed consistent improvements over conventional VGG16 transfer learning.

The comparison with other PEFT strategies showed that CoLoRA provides a competitive alternative to existing CNN adaptation methods, although it is not necessarily the least expensive method during training or the best-performing method in every setting. \change{The central contribution of CoLoRA is the integration of structured pointwise--depthwise kernel adaptation, exact fusion of the learned
updates into the pretrained convolutional weights, and an optional merge--reset mechanism that allows structured updates to accumulate
during training. Thus, the novelty of CoLoRA does not lie in low-rank adaptation, depthwise-separable convolutions, or weight merging
individually, but in combining these elements into a unified convolution-specific PEFT formulation. This formulation does not require
an explicit per-update rank or scaling-factor hyperparameter, although repeated merge--reset stages can progressively increase the capacity of the cumulative update.} The layer-placement experiments further showed that adaptation performance and computational cost depend strongly on where CoLoRA is introduced, with deeper convolutional blocks generally providing a favorable efficiency–performance trade-off.

The non-medical experiments further support the applicability of CoLoRA beyond medical-image classification. On CIFAR-100, CoLoRA applied to blocks 2--5 achieved the highest mean AUC, Top--1 accuracy, and Top--5 accuracy among the evaluated methods. The same CoLoRA placement also achieved the highest mean values for all three metrics in the controlled ImageNet--R evaluation,
with higher mean values than Transfer Learning and Conv-LoRA under the evaluated settings. These descriptive repeated-run results provide consistent evidence across the two non-medical benchmarks that adapting multiple deeper convolutional blocks can be effective when transferring the pretrained representation to a new image domain. However, additional experiments across datasets and architectures are necessary
to determine how generally this behavior extends to other tasks and convolutional backbones.

At the same time, the GPU-memory analysis showed that parameter reduction does not translate directly into proportional training-memory savings: memory consumption depends strongly on the placement of the adapted layers, particularly because early convolutional blocks operate on higher-resolution feature maps.

Overall, CoLoRA provides a parameter-efficient and deployment-efficient alternative to full fine-tuning for convolutional architectures, offering a structured and mergeable adaptation mechanism whose predictive performance, training cost, and memory footprint can be controlled through the selection of the adapted convolutional layers.

\section{Future Work}
\label{sec:future}

Future work will evaluate CoLoRA on larger and more heterogeneous datasets and additional modalities, including fundus, dermoscopic, and radiographic images. We will also implement this general proposal to 3D and spatiotemporal tasks such as time series, volumetric OCT/CT/MRI, and segmentation. Since CoLoRA does not use an explicit per-update LoRA rank hyperparameter, future work will focus on adaptive layer selection, systematic comparisons of merge schedules (including no intermediate merging, every-epoch merging, and merging every $k$ epochs), and capacity-control strategies based on kernel size, channel count, network depth, and cumulative effective rank. Beyond the current backbones, we plan to benchmark CoLoRA in additional convolutional architectures, including EfficientNet \cite{Efficientnetv2}, ConvNeXt \cite{Convnext}, and UNet-style segmentation models \cite{UNet}. These directions will further evaluate the generality and practical applicability of CoLoRA across domains, scales, architectures, and dimensionalities.

\section*{Declaration of generative AI and AI-assisted technologies in the manuscript preparation process.}
\label{sec:IA}

Statement: During the preparation of this work, the author(s) used ChatGPT in order to improve its organization and readability. After using this tool/service, the authors reviewed and edited the content as needed and take full responsibility for the content of the published article. 

\section*{Acknowledgment}
Work supported in part by SECIHTI, Mexico (Grants CB-A1-43858 and ApoyoLN-2025-C-3)

\end{document}